\pdfoutput=1

\documentclass[11pt]{article}

\usepackage[final]{acl}

\usepackage{times}
\usepackage{latexsym}

\usepackage[T1]{fontenc}

\usepackage[utf8]{inputenc}

\usepackage{microtype}

\usepackage{inconsolata}

\usepackage{graphicx}

\usepackage{amsmath}
\usepackage{multirow}
\usepackage{amssymb}
\usepackage{xspace}
\usepackage{enumitem}
\usepackage{tcolorbox}
\usepackage{booktabs}
\usepackage{multicol}
\usepackage{cuted}
\usepackage{tabularx}
\usepackage{hyperref}
\usepackage{listings}
\usepackage{xcolor}
\definecolor{codegray}{rgb}{0.5, 0.5, 0.5}
\definecolor{codepurple}{rgb}{0.5, 0, 0.5}
\lstdefinestyle{mystyle}{
    keywordstyle= \color{ blue!70},			
    commentstyle= \color{red!50!green!50!blue!50},		
    numberstyle=\tiny\color{codegray},		
    stringstyle=\color{codepurple},
    basicstyle=\Inconsolata\footnotesize,
    breakatwhitespace=false,         
    breaklines=true,		
    captionpos=b,                    
    keepspaces=true,                 
    numbers=left,		
    numbersep=5pt,                  
    showspaces=false,                
    showstringspaces=false,		
    showtabs=false,                  
    tabsize=2,
    frame=shadowbox,	
}

\newcommand{\eg}{\emph{e.g.}\xspace}

\newcommand{\ignore}[1]{}

\usepackage{lipsum}
\usepackage{environ}
\NewEnviron{SMALL}{%
    \scalebox{1}{$\BODY$} 
} 
\newtheorem{definition}{Definition}

\title{Instructions for *ACL Proceedings}

\author{First Author \\
  Affiliation / Address line 1 \\
  Affiliation / Address line 2 \\
  Affiliation / Address line 3 \\
  \texttt{email@domain} \\\And
  Second Author \\
  Affiliation / Address line 1 \\
  Affiliation / Address line 2 \\
  Affiliation / Address line 3 \\
  \texttt{email@domain} \\}

\title{2D-DPO: Scaling Direct Preference Optimization\\ with 2-Dimensional Supervision}

\author{
Shilong Li$^{*\dag}$,
Yancheng He$^{*}$,
Hui Huang$^{*}$,
Xingyuan Bu$^{*\ddag}$,
Jiaheng Liu,
\\
{\bf Hangyu Guo, Weixun Wang, Jihao Gu, Wenbo Su, Bo Zheng} \\
Alibaba Group\ \ \ \\
{\tt zhuli.lsl@taobao.com, xingyuanbu@gmail.com}
}

\begin{document}
\maketitle
\let\thefootnote\relax\footnotetext{$*$ Equal contribution. $\ddag$ Corresponding Author.}
\let\thefootnote\relax\footnotetext{$\dag$ Work done during an internship at Alibaba Group.}
\begin{abstract}
Recent advancements in Direct Preference Optimization (DPO) have significantly enhanced the alignment of Large Language Models (LLMs) with human preferences, owing to its simplicity and effectiveness.  However, existing methods typically optimize a scalar score or ranking reward, thereby overlooking the multi-dimensional nature of human preferences.
In this work, we propose to extend the preference of DPO to two dimensions: \textbf{segments} and \textbf{aspects}. We first introduce a 2D supervision dataset called \textbf{HelpSteer-2D}. For the segment dimension, we divide the response into sentences and assign scores to each segment. For the aspect dimension, we meticulously design several criteria covering the response quality rubrics. With the 2-dimensional signals as feedback, we develop a \textbf{2D-DPO} framework, decomposing the overall objective into multi-segment and multi-aspect objectives. 
Extensive experiments on popular benchmarks demonstrate that 2D-DPO performs better than methods that optimize for scalar or 1-dimensional preferences\footnote{Codes and datasets are anonymously at \url{https://anonymous.4open.science/r/2D-DPO-56E4/} and will be released to the public once accepted to promote related research.}.
\end{abstract}
\section{Introduction}

Recent advancements in Large Language Models (LLMs) have shown impressive performance across a wide range of tasks~\cite{zhao2023survey,bai2024mt,wu2024conceptmath,li2024graphreader}. A pivotal component in LLM training is Reinforcement Learning from Human Feedback (RLHF) ~\cite{ouyang2022training, bai2022training}, which aligns LLMs with human preferences.
However, due to its complexity, traditional RLHF often leads to challenges such as training instability and reward collapse ~\cite{wolf2023fundamental, song2023reward}.

Direct Preference Optimization (DPO)~\cite{rafailov2023direct}, as a simpler and more effective alternative, has gained considerable attention due to its ability to bypass the need for explicitly fitting a reward model~\cite{meng2024simpo, ethayarajh2024kto}.
However, most existing DPO-style approaches rely on scalar scores or rankings and ignore the multi-dimensional nature of human preferences, resulting in inefficient and imprecise optimization.
For instance, a response may be deemed satisfactory under one aspect such as \texttt{correctness}, but falls short in another such as \texttt{clarity}. Moreover, not all segments of a response should be treated uniformly; even in a preferred response, there may be segments of inferior quality. This underscores the need for a more nuanced approach that recognizes the multi-dimensionality of feedback and its critical impact on model training.

\begin{figure}[t]
    \centering
    \includegraphics[width=1\linewidth]{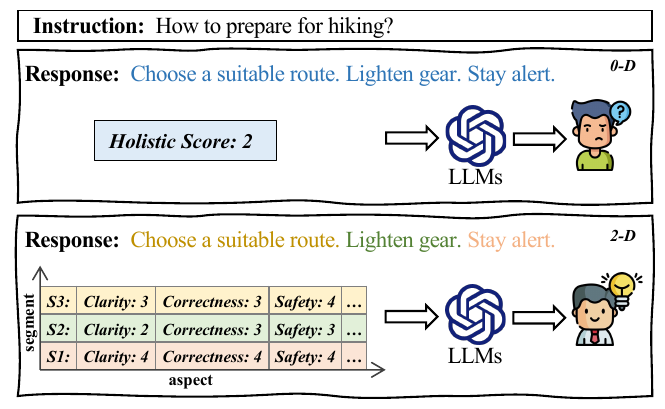}
    \caption{An illustrative comparison between vanilla DPO and 2D-DPO.}
    \label{fig: intro}
    \vspace{-1mm}
\end{figure}
 
In response, some recent works have attempted to leverage signals that are believed to reflect the importance of individual segments as reward scores~\cite{zeng2024token, chan2024dense, jiang2024bridging, chen2024improving}. However, these signals are often derived from statistical features such as edit distance or confidence estimation, which can introduce noise and lack interpretability. Other approaches incorporate multi-objective optimization to balance different aspects of human preferences~\cite{wu2024fine, guo2023beyond, cao2024drlc}. However, these methods mostly rely on Proximal Policy Optimization (PPO) \cite{schulman2017proximal}, which is prone to instability during training. Furthermore, these efforts only extend preference optimization from 0-dimensional (scalar reward) to 1-dimensional (aspect/segment) supervision, which remains insufficient for capturing the complexity of real-world human preferences.

To better address the intricacy of human preferences, we propose \textbf{2D-DPO}, a novel direct alignment strategy that enables 2-dimensional (2D) fine-grained optimization. 
Our core idea is to scale supervision signals across two dimensions: \textbf{segments} and \textbf{aspects}. To this end, we first construct a preference dataset called HelpSteer-2D, where each sample is annotated with a 2-dimensional score matrix evaluating each segment across multiple aspects. These signals are derived from a robust model guided by a set of stringent principles, ensuring the generation of highly accurate and interpretable supervision signals. 
Building on this, we propose a novel approach to achieve 2-dimensional direct preference alignment.
Experimental results on three public benchmarks demonstrate that 2D-DPO significantly outperforms previous methods.
In summary, our main contributions are threefold:
\begin{itemize}[leftmargin=4mm]
\item We introduce a novel 2-dimensional preference alignment method, 2D-DPO, which scales supervision signals across both segments and aspects to better align with human preferences.
\item We develop a high-quality, fine-grained preference dataset, HelpSteer-2D, which will be released to the community for future research.
\item Extensive experiments show that 2D-DPO delivers superior performance in aligning with human preferences compared to prior approaches.
\end{itemize}

\section{Related Work}
\subsection{Preference Optimization}

Large language models (LLMs) have advanced rapidly, with reinforcement learning from human feedback (RLHF) commonly used to align LLMs with human preferences~\citep{ziegler2019fine, stiennon2020learning, ouyang2022training, bai2022training, liu2024iterative, liu2024dream, peng2023gaia, feng2022beyond}. However, traditional RLHF methods face challenges like instability and high resource demands~\cite{wolf2023fundamental, song2023reward}, prompting the search for simpler alternatives. 
One such representative approach is Direct Preference Optimization (DPO)~\cite{rafailov2023direct}, which optimizes alignment without explicit reward modeling, offering simplicity and stability. Building on DPO, IPO~\cite{azar2024general} adds a regularization term to alleviate overfitting. KTO~\cite{ethayarajh2024kto} only requires a binary signal of whether an output is desirable or undesirable for an input to align LLMs, simplifying the data acquisition process. ORPO~\cite{hong2024reference} simplifies training with odds ratio-based penalties, and SimPO~\cite{meng2024simpo} improves efficiency by using average log probability as an implicit reward.

\subsection{Token-level Preference Optimization}

The response-level rewards in naive PPO and DPO often lack token-level details.
To address this, researchers have explored fine-grained supervision signals in three ways:
(1) Human annotation: Methods like PRM~\cite{lightman2023let} and FGRLHF~\cite{wu2024fine}  involve human annotators labeling each segment of the response to generate fine-grained signals.
(2) LLM annotation: To reduce the cost of human labeling, stronger LLMs are used to generate preference pairs with minimal edits ~\cite{guo2023beyond,chen2024improving,yoon2024tlcr,jiang2024bridging} or to identify positive and negative response segments~\citep{cao2024drlc}.
(3) Internal signal: Some works use the model's internal information as reward signals, such as using attention scores for token rewards in ABC~\cite{chan2024dense} or decomposing DPO’s response-level rewards into token-level signals in SePO~\cite{yang2024selective, rafailov2024r}. TDPO~\cite{zeng2024token} achieves token-level alignment by controlling the KL divergence for each token.

\subsection{Multi-objective Preference Optimization}

Human preferences are often complex, diverse, and even contradictory, making single-dimensional training insufficient. To address this, some studies align LLMs with multiple objectives by either training separate reward models and averaging their outputs~\cite{pan2023rewards,ji2024beavertails,rame2024rewarded,de2024reinforcement, wang2024arithmetic}. However, this approach demands significant computational resources. In contrast, MODPO~\cite{zhou2024beyond} offers a simpler, reinforcement learning-free method for optimizing multiple objectives. RiC~\cite{yang2024rewards} and CPO~\cite{guo2024controllable} focus on integrating multiple reward values for controllable generation.
\section{Approach}

\begin{figure*}[t]
    \centering
    \includegraphics[width=1\linewidth]{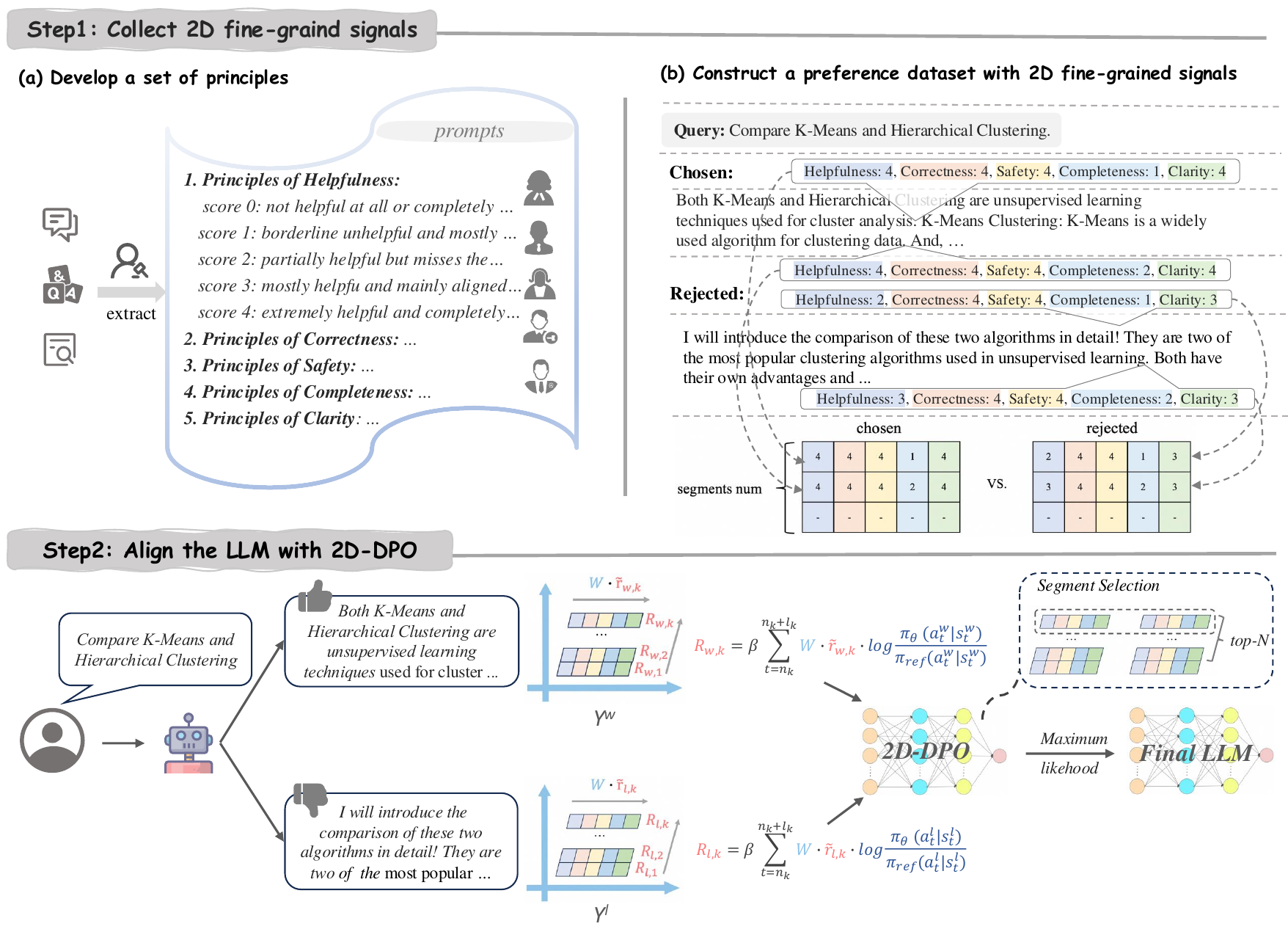}
    \vspace{-5mm}
    \caption{Illustration of our proposed 2D-DPO. Firstly, we develop principles for preference annotation on different aspects, and collect scores across different segments and aspects for pairwised responses, leading to 2-dimensional signals. Secondly, we apply 2D-DPO on the constructed signals with decomposed training objective.}
    \label{fig: ourmethod}
    \vspace{-3mm}
\end{figure*}

In this section, we provide a detailed introduction to our 2D-DPO approach for aligning LLMs.
We first describe the construction of the 2D preference dataset in Section \ref{app: dataset}. Then, we present an enhanced 2D-DPO formulation that integrates the 2D supervision signals into the alignment process in Section \ref{app: formula}.
The complete pipeline of our method is illustrated in  Figure~\ref{fig: ourmethod}.

\subsection{Preference Dataset with 2D Signal}
\label{app: dataset}
In general, a preference optimization dataset, denoted as $\mathcal{D} = \{x^i, y_w^i, y_l^i\}_{i=1}^N$, comprises prompts $x$ along with a chosen response $y_w$ and a rejected response $y_l$, where $y_w$ is of higher quality compared to $y_l$. Such datasets are commonly used to train reward models (\eg, PPO) or directly for model alignment (\eg, DPO). However, differentiating between the chosen and rejected responses based on a scalar score is often coarse and imprecise. The quality of responses can vary significantly across different evaluation aspects, and even a chosen response might contain segments of low quality, while a rejected response could include segments of high quality. Therefore, relying solely on a scalar score for optimization may restrict the model's ability to effectively align with human preferences.

To address this issue, we propose a fine-grained scoring approach that decomposes the scalar scores of model responses to segment-level and aspect-level. The first step is to divide the response into segments, and the choice of segment length is crucial for ensuring the effectiveness of the fine-grained optimization. Segments that are too long cannot resolve the aforementioned coarse scoring issues, while segments that are too short pose difficulties for accurate assessment. Therefore, we choose the sentence as the scoring unit, which can strike a balance between scoring accuracy and the clarity of segment preferences.

After segmenting responses based on typical sentence-ending punctuations, we employ GPT-4 to perform aspect-level scoring. Following HelpSteer2~\cite{wang2024helpsteer2}, we annotate the preference data across five key aspects: \texttt{Helpfulness}, \texttt{Correctness}, \texttt{Safety}, \texttt{Completeness}, \texttt{Clarity}. The first three aspects are independent of different sentences. The aspect of \texttt{Completeness} generally increases as responses become more comprehensive, while \texttt{Clarity} tends to decrease as responses grow longer and more redundant. To ensure the integrity of our annotations, we use separate prompts for each aspect to prevent any cross-influence among them. For the details of the annotation process, please refer to Appendix ~\ref{app: annotation}.

Finally, the constructed dataset is as follows:
\begin{align}
& \mathcal{D} = \{x^i, \boldsymbol{y}^i_w, \boldsymbol{y}^i_l\}_{i=1}^N  \nonumber \\
& \boldsymbol{y}^i_w = \{y^i_{w,k}, \{r^i_{w,k,j}\}_{j=1}^{A}\}_{k=1}^{S^i_w}, \nonumber \\
& \boldsymbol{y}^i_l = \{y^i_{l,k}, \{r^i_{l,k,j}\}_{j=1}^{A}\}_{k=1}^{S^i_l}.
\end{align}

In the dataset $\mathcal{D}$, $x^i$ represents the $i$-th prompt, while $\boldsymbol{y}^i_w$ and $\boldsymbol{y}^i_l$ represent the 2D annotations for the chosen and rejected response, respectively. Each 2D annotation includes $S^i$ text segments, indexed by $k$, denoted as $y^i_k$. The scores for these segments are denoted by $r^i_{k,j}$, where $j$ indicates the index across the aforementioned five aspects.

\subsection{2D-DPO}
\label{app: formula}
While the construction process of 2D signals is straightforward, integrating them effectively into the alignment process presents significant challenges. Previous approaches mostly utilize these signals as a scalar reward by weighted summation, which is insufficient for enabling the model to distinguish between varying quality across different dimensions. To address this issue, we propose a novel alignment method called 2D-DPO.
\paragraph{Vanilla DPO.}
Direct Preference Optimization (DPO) \cite{rafailov2023direct}, as one of the most popular alignment methods, proposes a direct optimization objective that satisfies the optimal preference policy without using a reward model:
\begin{align}
& \mathcal{L}_{DPO}(\pi_\theta; \pi_{ref}) = \nonumber \\
& -\mathbb{E}_{\left(x, y^w, y^l \right) \sim \mathcal{D}} \left[ \log \sigma \left( \beta \log \frac{\pi_\theta(y_w \mid x)}{\pi_{ref}(y_w \mid x)} \right. \right. \nonumber \\
& \left. \left. - \beta \log \frac{\pi_\theta(y_l \mid x)}{\pi_{ref}(y_l \mid x)} \right) \right], 
\end{align}
\noindent where $\pi_\theta$ and $\pi_{ref}$ represent the policy model and the reference model, respectively. DPO can fundamentally be viewed as a multi-armed bandit problem, where the model's entire response is treated as a single arm.
According to \citealt{rafailov2023direct}, in the token-level Markov Decision Process (MDP), the language model's Bradley-Terry preference model can be expressed in the following form:
\begin{align}
&
p^*(\tau^w \succeq \tau^l) = 
\label{eq: bt} \\
&
\frac{\exp \left( \sum_{i=1}^N r(\mathbf{s}_i^w, \mathbf{a}_i^w) \right)}{\exp \left( \sum_{i=1}^N r(\mathbf{s}_i^w, \mathbf{a}_i^w) \right) + \exp \left( \sum_{i=1}^M r(\mathbf{s}_i^l, \mathbf{a}_i^l) \right)},
\nonumber
\end{align}
\noindent where $\tau^w$ and $\tau^l$ represent the winning and losing trajectories, respectively. In this context, $\mathbf{a}$ represents the next generated token, and $\mathbf{s}$ denotes the current state, consisting of the prompt along with all previously generated tokens.

Subsequently, based on the fixed point solution under the general maximum entropy RL setting \citep{Ziebart2010ModelingPA, Levine2018ReinforcementLA}, \citeauthor{rafailov2024r} derived the form of DPO in token-level MDP:
\begin{align}
& \mathcal{L}_{DPO}(\pi_\theta; \pi_{ref}) = \\
& -\mathbb{E} \left[ \log \sigma \left( \beta \sum_{t=0}^{N-1} \log \frac{\pi_\theta(a_w^t \mid s_w^t)}{\pi_{ref}(a_w^t \mid s_w^t)} \right. \right. \nonumber \\
& \left. \left. \quad \quad \quad \quad - \beta \sum_{t=0}^{M-1} \log \frac{\pi_\theta(a_l^t \mid s_l^t)}{\pi_{ref}(a_l^t \mid s_l^t)} \right) \right].  \nonumber
\end{align}
\indent In the above process, \citeauthor{rafailov2024r} combined \citeauthor{Ng1999PolicyIU}'s definition of equivalence between two reward functions through a potential function and concluded that $\beta \log \frac{\pi^*_{\theta}(\mathbf{a}_t \mid \mathbf{s}_t)}{\pi_{ref}(\mathbf{a}_t \mid \mathbf{s}_t)}$ and $r(\mathbf{s}_t, \mathbf{a}_t)$ can equivalently yield the same optimal policy. Furthermore, $\beta \log \frac{\pi^*_{\theta}(\mathbf{a}_t \mid \mathbf{s}_t)}{\pi_{ref}(\mathbf{a}_t \mid \mathbf{s}_t)}$ is precisely the optimal advantage function $A^*(\mathbf{s}_t, \mathbf{a}_t)$. For a detailed derivation, please refer to Appendix \ref{app: tdpo} and \ref{app: optimal_adv}.

\paragraph{2D-DPO.}
With the above conclusions, 2D supervision signals can be conveniently integrated into the alignment process. We achieve the integration by using the signals to calibrate the token-level advantage function $A(\mathbf{s}_t, \mathbf{a}_t)$ for different segments. This approach equips the model with a direct perception of fine-grained preferences, avoiding the ambiguity of holistic rewards.

Specifically, we use the regularized fine-grained reward $r$ as a coefficient, incorporating it into the token-level advantage function to obtain $\beta r \log \frac{\pi_{\theta}(\mathbf{a}_t \mid \mathbf{s}_t)}{\pi_{ref}(\mathbf{a}_t \mid \mathbf{s}_t)}$, which integrates the fine-grained signals. In practice, this is equivalent to adjusting $\beta$ within the original $A(\mathbf{s}_t, \mathbf{a}_t)$. We provide complete proof of its feasibility in Appendix \ref{app: change_beta}.

The token-level DPO incorporating fine-grained signals is formalized as follows:
\begin{small}
\begin{align}
& \mathcal{L}(\pi_{\theta},D) = 
\nonumber 
\\
& -\mathbb{E}_{(\tau_w,\tau_l)\sim D} \log \sigma ( \beta \sum_{k=0}^{S_w-1} \sum_{t=n_{k}}^{n_{k}+l_k} r_{w,k} \log \frac{\pi_{\theta}(\mathbf{a}_t^w|\mathbf{s}_t^w)}{\pi_{ref}(\mathbf{a}_t^w|\mathbf{s}_t^w)} 
\nonumber 
\\
& -\beta \sum_{k=0}^{S_l-1} \sum_{t=n_{k}}^{n_{k}+l_k} r_{l,k} \log \frac{\pi_{\theta}(\mathbf{a}_t^l|\mathbf{s}_t^l)}{\pi_{ref}(\mathbf{a}_t^l|\mathbf{s}_t^l)}),
\label{eq: rdpo}
\end{align}
\end{small}
\noindent where $n_k$ represents the first token of the $k$-th segment and $l_k$ denotes the length of the $k$-th segment.

For handling multiple aspects, we use a classic weighted approach for integration, that is, $r_{w,k} = \mathbf{W} \mathbf{\vec{r}}_{w,k}$, where $\mathbf{W}$ represents the weights that sum up to 1, which reflect the importance of each aspect during the alignment process, and $\mathbf{\vec{r}}_{w,k} = \{r_{w,k,j}\}_{j=1}^A$.

\paragraph{Segment Selection.}
The number of segments in the chosen and rejected responses may differ significantly, and typically only the segments with an impact on response preference need attention. Therefore, we select the top-$N$ highest-scoring segments from the chosen response and the top-$N$ lowest-scoring segments from the rejected response, where $N = \min(S_w, S_l)$, further enhances the efficiency of model alignment training. Additionally, we group segments in pairs to provide clearer contrast during alignment, making it easier for the model to learn fine-grained differences between the chosen and rejected responses. These segments are paired to form $N$ BT models. The feasibility of this rearrangement is based on the fact that the loss for a single-segment BT model can be treated as setting the $\beta_t$ of other parts to 0, as demonstrated in Appendix \ref{app: change_beta}. Thus, we obtain the token-level DPO formula incorporating fine-grained signals:

\begin{small}
\begin{align}
    & \mathcal{L}_{group}(\pi_{\theta},D) = 
    \nonumber \\
    & -\mathbb{E}_{(\tau_w,\tau_l)\sim D} \left[ \sum_{k=0}^{N-1} \log \sigma \left( \beta \sum_{t=n_{k}}^{n_{k}+l_k} r_{w,k} \log \frac{\pi_{\theta}(\textbf{a}_t^w|\textbf{s}_t^w)}{\pi_{ref}(\textbf{a}_t^w|\textbf{s}_t^w)}
    \right. \right. \nonumber \\
    & \left. \left. -\beta \sum_{t=n_{k}}^{n_{k}+l_k} r_{l,k} \log \frac{\pi_{\theta}(\textbf{a}_t^l|\textbf{s}_t^l)}{\pi_{ref}(\textbf{a}_t^l|\textbf{s}_t^l)} \right) \right].
\end{align}
\label{eq:main}
\end{small}

As a result, we've formulated the definitive objective of 2D-DPO. This training objective allows for the direct integration of 2-D supervision signals into the alignment process, enabling LLMs to discern the different aspects lying in different segments in the responses, thereby promoting better alignment with human preferences.

\section{Experiments}
\subsection{Set-up}

\setlength{\tabcolsep}{3pt}
\begin{table*}[]
\resizebox{1.0\textwidth}{!}{
\begin{tabular}{c|l|ccc|ccc|ccc}
\toprule
\multirow{2}{*}{\textbf{Policy Model}} & \multirow{2}{*}{\textbf{Methods}} & \multicolumn{3}{c|}{\textbf{Arena-Hard}} & \multicolumn{3}{c|}{\textbf{AlpacaEval 2.0}} & \multicolumn{3}{c}{\textbf{MT-Bench}}  \\  &  & \textbf{WR (\%)} & \textbf{Avg. Len} & \textbf{95\% CI} & \textbf{LC (\%)} & \textbf{WR (\%)} & \textbf{Avg. Len} & \textbf{Turn 1} & \textbf{Turn 2} & \textbf{Avg. Score} \\ 
\midrule
\multirow{9}{*}{\begin{tabular}[c]{@{}c@{}}Qwen2-7B\\ -Instruct\end{tabular}}  & Base & 25.10             & 583 & (-2.1, 2.0) & 30.68  & 28.32  & 1862  & 8.01 & 6.61 & 7.31 \\ \cline{2-11} 
& + DPO  & 29.40 & 578 & (-1.8, 1.9) & 29.07 & 26.83 & 1996 & 8.11 & 6.45 & 7.28 \\
& + IPO  & 26.50 & 556 & (-2.2, 2.2) & 28.70 & 26.58 & 1940 & 7.90 & 6.53 & 7.21 \\
& + KTO  & 26.10 & 518 & (-2.1, 2.5) & 26.46 & 23.00 & 1730 & 8.11 & 6.40 & 7.26 \\
& + ORPO & 25.40 & 573 & (-2.2, 1.8) & 28.58 & 27.70 & 1936 & 8.09 & 6.52 & 7.31 \\ 
& + SimPO & 29.00 & 539 & (-1.9, 2.4) & 29.94 & 27.70 & 1904 & 8.06 & 6.50 & 7.28 \\
\cline{2-11}
& + TDPO & 25.90 & 564 & (-2.0, 2.4)& 29.81 & 27.33 & 1896 & 8.05 & 6.46 & 7.26 \\
\cline{2-11}
& + 1D-DPO & 29.80 & 574 & (-2.4, 2.2) & 31.07 & 28.70 & 1951 & 8.13 & 6.48 & 7.31  \\
& + \textbf{2D-DPO} & \textbf{30.30} & 586 & (-1.9, 2.4) & \textbf{31.51} & 28.94 & 1994 & \textbf{8.18} & 6.68 & \textbf{7.43} \\ 
\midrule
\multirow{9}{*}{\begin{tabular}[c]{@{}c@{}}Llama3-8B\\ -Instruct\end{tabular}} & Base                      & 25.40 & 599 & (-2.2, 2.4) & 27.08 & 26.96 & 1959 & 7.66 & 6.84 & 7.25 \\ \cline{2-11} 
& + DPO & 25.90 & 567 & (-1.9, 2.2) & 31.68 & 30.31 & 1883 & 7.64 & 6.60 & 7.20 \\
& + IPO & 24.80 & 548 & (-2.2, 1.8) & 29.69 & 28.57 & 1891 & 7.73 & 6.75 & 7.24 \\
& + KTO & 25.20 & 507 & (-1.9, 2.4) & 27.95 & 27.08 & 1835 & 7.65 & 6.65 & 7.15 \\
& + ORPO & 25.60 & 537 & (-1.7, 1.9) & 29.19 & 28.57 & 1892 & 7.75 & 6.68 & 7.22 \\
& + SimPO  & 26.30  & 552 & (-1.8, 2.2) & 31.55  & 30.19 & 1879  & 7.96 & 6.70  & 7.33  \\ \cline{2-11} 
& + TDPO & 23.40 & 566 & (-1.7, 1.8) & 28.57 & 26.96 & 1881 & 7.95 & 6.80 & 7.38 \\ \cline{2-11} 
& + 1D-DPO  & 26.70 & 563  & (-2.3, 1.8)  & 31.78 & 30.19 & 1893 & 7.98 & 6.74 & 7.34 \\
& + \textbf{2D-DPO} & \textbf{27.00} & 554 & (-1.9, 2.0) & \textbf{32.06} & \textbf{30.56} & 1884 & \textbf{8.04}  & 6.84  & \textbf{7.44} \\ 
\bottomrule
\end{tabular}}
\caption{Experiment results of different preference optimization methods on instruction-following benchmarks. We report the results on each benchmark based on their recommended metrics.}
\label{tab: main}
\end{table*}

\paragraph{Benchmark.} Our method has been tested on three wildly recognized instruction-following benchmarks: Arena-Hard \cite{li2024crowdsourced}, AlpacaEval 2.0 \cite{dubois2024length} and MT-Bench \cite{zheng2023judging}. Each benchmark comprises a diverse set of queries, and the answers are evaluated under the framework of LLM-as-a-Judge \cite{zheng2023judging}. We use \texttt{gpt-4-turbo-2024-04-09}\footnote{\url{https://platform.openai.com/docs/models/gpt-4-turbo-and-gpt-4}} as the judge model, and scores are reported following each benchmark's protocol.

\paragraph{Model.} Our method is validated on two models, Qwen2-7B-Instruct \cite{qwen2} and Llama-3-8B-Instruct \cite{llama3modelcard}. It deserves to be noticed that both models have undergone extensive instruction-tuning processes, therefore we directly perform preference optimization.

\paragraph{Baseline.} We mainly compare our method with 0-dimentional preference optimization methods:

\begin{itemize}
    \item DPO \cite{rafailov2023direct}. This method leverages a mapping between reward functions and optimal policies to optimize the preference with a single stage of policy training.
    \item IPO \cite{azar2024general}. This method propose a theoretically grounded approach method to replace pairwise preferences in DPO with pointwise rewards.
    \item KTO \cite{ethayarajh2024kto}. This method proposes to directly maximize the utility of generations from non-paired data.
    \item ORPO \cite{hong2024reference}. This method leverages a reference model-free monolithic odds ratio for contrasting favored and disfavored styles during SFT stage.
    \item SimPO \cite{meng2024simpo}. This method proposes to use the average log probability of a sequence as the implicit reward, which eliminates the need for a reference model.
\end{itemize}

We also compare our method with the following 1-dimensional preference optimization method:

\begin{itemize}
    \item TDPO \cite{zeng2024token}. This method proposes to control the  KL divergence constraints for each token, aiming to strike a better balance between alignment and diversity.
\end{itemize}

Our method is evaluated under two configurations — 1D-DPO and 2D-DPO. 1D-DPO only incorporates a single aspect (\texttt{helpfulness}) as signals while 2D-DPO uses all five aspects.

\paragraph{Training.} We perform preference optimization based on HelpSteer-2D, which is constructed based on HelpSteer2 \cite{wang2024helpsteer2}. We leverage \texttt{gpt-4o-2024-05-13}\footnote{\url{https://platform.openai.com/docs/models/gpt-4o}} to generate 2-dimensional scores which align with our requirements. 

To make a fair comparison, all methods are combined with SFT loss with a coefficient of 0.1 except for ORPO\footnote{ORPO is conducted with SFT loss originally.}. The other hyper-parameters are tuned to achieve an optimal performance for each method. Please refer to Appendix \ref{app:hyperparams} for more details.

\subsection{Main Results}
The primary results are shown in Table \ref{tab: main}. As can be seen, our proposed 2D-DPO outperforms existing methods across all three benchmarks, verifying the significance of 2-dimensional supervision in preference optimization. While previous methods mostly treat different segments uniformly with a singular scoring criterion, leading to insufficient supervision, this work scales the feedback to both aspect-level and segment-level, thereby improving the performance. Notice all methods are conducted on the same group of queries and responses, and our method does not require an aditional training stage or extra computation overhead, therefore our method exploits the utility of direct preference optimization with minimal expense.

Comparing the averaged length on Arena-Hard and AlpacaEval 2.0, we also notice that our method does not lead to more verbose responses. This demonstrates that 2-dimensional supervision is helpful for mitigating the reward hacking issue \cite{singhal2023long}. While more fine-grained supervision is provided for preference pairs, the model would not unanimously favor more verbose responses, achieving more accurate alignment.

In comparison to 2D-DPO, the performance of 1D-DPO, which is only scaled with segment-level signals, shows a noticeable decline, thereby demonstrating the gains achieved by incorporating aspect-level signals. Nevertheless, among various alignment methods, 1D-DPO remains highly competitive, outperforming all other methods except for 2D-DPO across different evaluation metrics, proving the effectiveness of the additional supervisory signals introduced at the segment level.

Furthermore, it deserves to be noted that TDPO also underperforms both 1D-DPO and 2D-DPO, which can be traced back to the design of training objective. Despite the loss in TDPO is re-assigned to each token based on KL-Divergence, the temperature for each token is not adjusted appropriately, resulting in a coarse optimization process. In contrast, our method meticulously adjust the temperature for each segment, aligning the update scale with the segment's importance across various criteria, which attribute to our superior performance.

\subsection{Detailed Analysis}

To further show the effectiveness of 2D-DPO, we conduct ablation studies and delve into a detailed analysis of the model's performance. Additionally, we present a case study in Appendix~\ref{contro_train}.

\paragraph{The Influence of $\beta$.}

\begin{table*}[!t]
\center
\resizebox{0.87\textwidth}{!}{
\begin{tabular}{c|c|ccc|ccc|ccc}
\toprule
\multirow{2}{*}{\textbf{Model}} & \multirow{2}{*}{\textbf{$\beta$}} & \multicolumn{3}{c|}{\textbf{Arena-Hard}}        & \multicolumn{3}{c|}{\textbf{AlpacaEval 2.0}} & \multicolumn{3}{c}{\textbf{MT-Bench}} \\
&  & \textbf{WR (\%)} & \textbf{Avg. Len} & \textbf{95\% CI} & \textbf{LC (\%)} & \textbf{WR (\%)} & \textbf{Avg. Len} & \textbf{Turn 1} & \textbf{Turn 2} & \textbf{Avg. Score} \\ 
\midrule
\multirow{5}{*}{} & 0.1 & 29.2 & 557 & (-2.2, 1.8) & 30.48 & 29.21 & 1961 & 8.00 & 6.60 & 7.30 \\
Qwen2-7B & 0.2 & \textbf{30.3} & 586 & (-1.9, 2.4) & \textbf{31.51} & 28.94 & 1994 & \textbf{8.18}   & \textbf{6.68}  & \textbf{7.43} \\
-Instruct & 0.5 & 27.4 & 578 & (-1.8, 1.9) & 29.62 & 29.21 & 1976 & 7.95 & 6.60 & 7.28 \\
+2D-DPO & 0.7 & 28.6 & 595 & (-1.6, 2.2) & 28.24 & 28.17 & 1997 & 7.83 & 6.65 & 7.24 \\
& 1.0 & 28.0 & 576 & (-2.0, 1.4) & 28.97 & 28.49 & 1981 & 7.77 & 6.65 & 7.21 \\ 
\bottomrule
\end{tabular}}
\caption{Experiment results on three benchmarks of 2D-DPO with different values of $\beta$.}
\vspace{-4mm}
\label{tab: beta}
\end{table*}

Table \ref{tab: beta} shows the results of 2D-DPO with different values of $\beta$ (temperature). As $\beta$ increases, we observe a consistent trend across three benchmarks: performance first rises and then falls. This is because a higher $\beta$ can amplify the divergence penalty within the RL optimization objective, thereby avoiding model degradation. However, an overly high $\beta$ would reduce the overall optimal ceiling and limit the potential gains from alignment \cite{ahrabian2024hitchhiker}.

\paragraph{Performance on Different Aspects.}

\begin{figure}[htbp]
\centering
\includegraphics[width=0.9\linewidth]{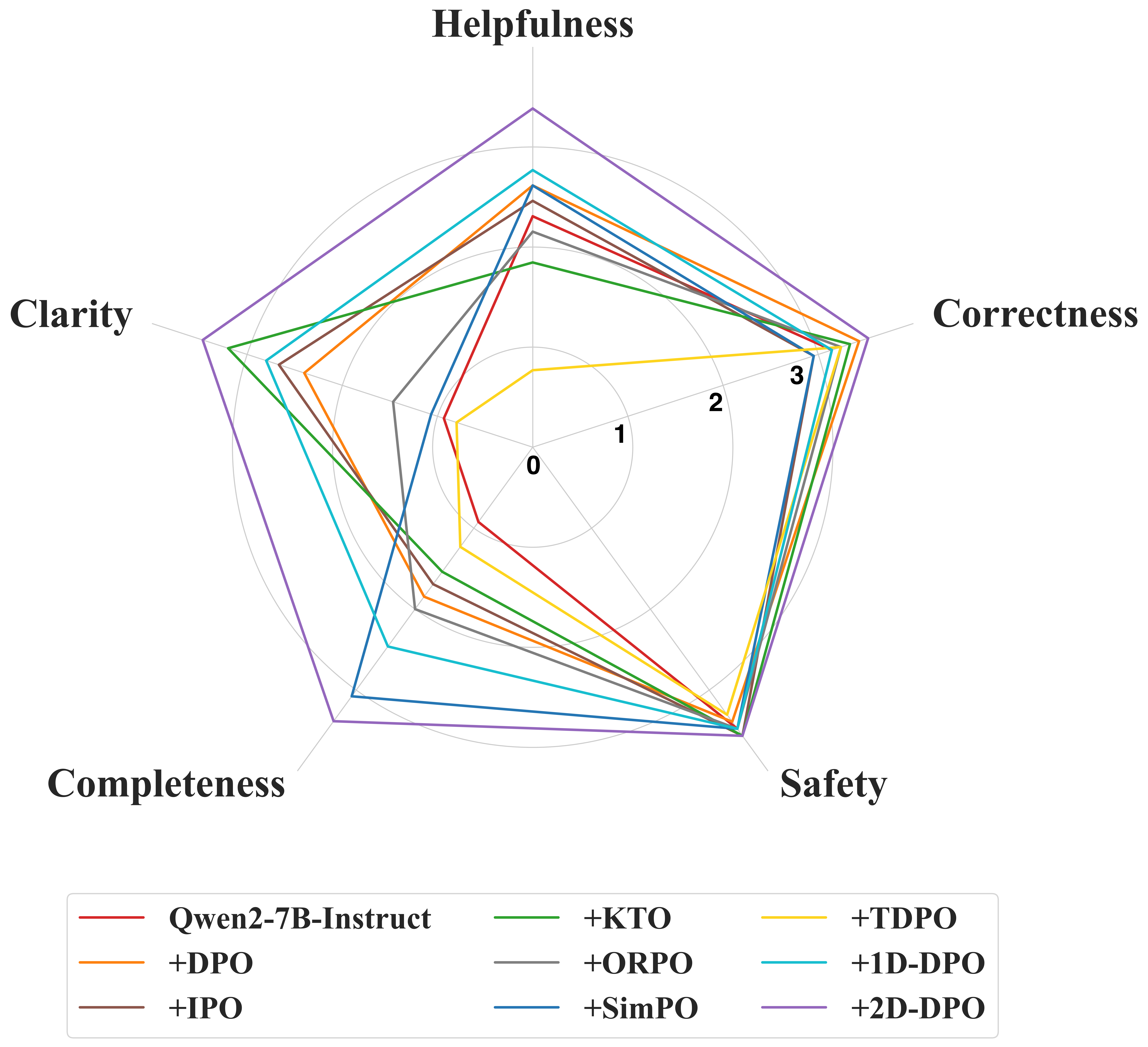}
\caption{The relative performance on different aspects of different alignment methods.} 
\label{fig: radar}
\vspace{-4mm}
\end{figure}

We evaluated the performance of the models aligned using different methods across various aspects. We selected AlpacaEval 2.0 \cite{dubois2024length} which offers diverse instructions as the query set, and obtained the responses of different models on this set as the evaluation targets. The evaluation prompt was consistent with the prompts presented in Section \ref{app: dataset}. For aspects that are independent among segments, we took the average score of all segments as the score for that response. For aspects that are not independent among segments (\texttt{completeness} and \texttt{clarity}), we select the score of the last segment. The average score of all responses is taken as the final result. As shown in Figure \ref{fig: radar}, our 2D-DPO can achieve the best results in all aspects, striking a balance between different rubrics of human preferences. 1D-DPO with only segment-level feedback underperforms, as response-level alignment still leads to coarse refinement. We also notice different methods exhibit minimal difference upon \texttt{safety} and \texttt{correctness}, which might be due to Qwen2-7B-Instruct already undergone alignment process on these aspects. For the other aspects that is not covered by the process, 2D-DPO can achieve more pronounced improvement.

\paragraph{Training Indicators.}
\begin{figure}[htbp]
\centering
\includegraphics[width=1\linewidth]{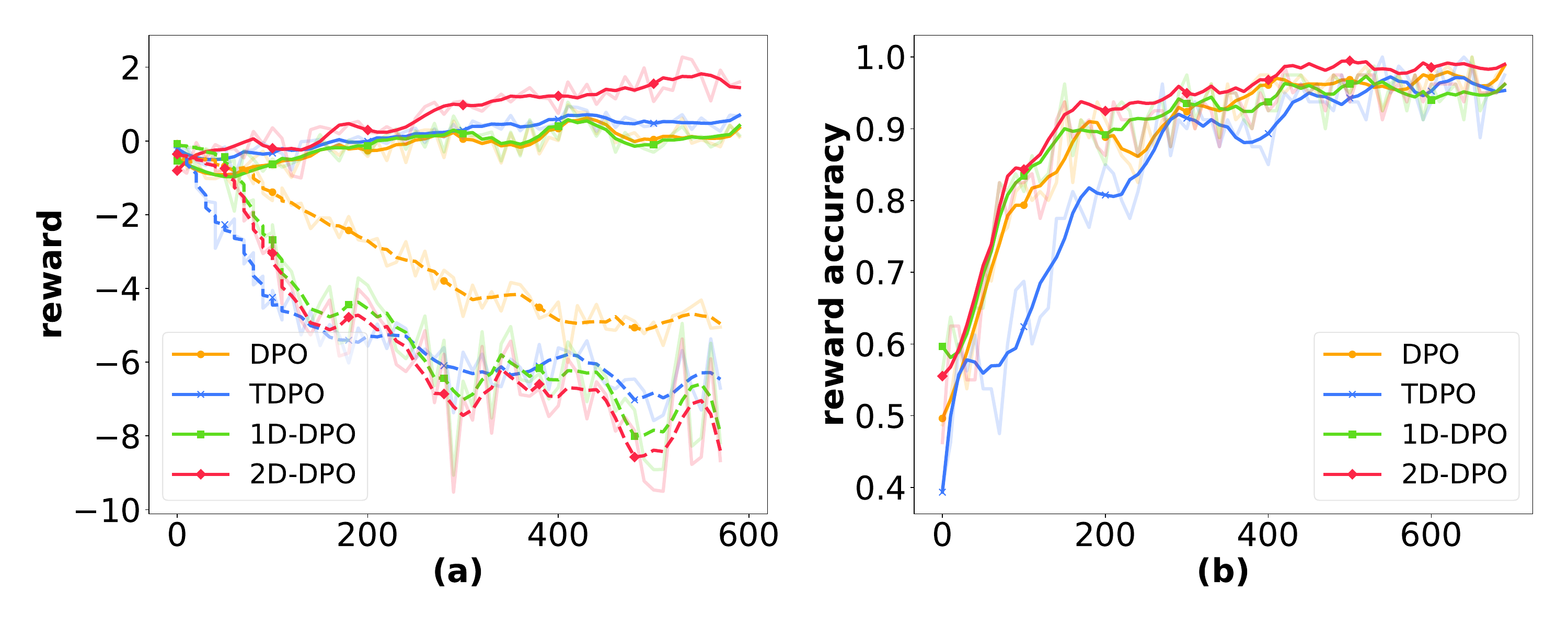}
\caption{The trends in reward scores and accuracy over training steps across DPO, TDPO, 1D-DPO, and 2D-DPO. (a) Rewards of preferred (solid lines) and dispreferred (dashed lines) responses. (b) Reward accuracy compared with preference annotation.} 
\label{fig: performance1}
\vspace{-4mm}
\end{figure}

\begin{figure}[htbp]
\centering
\includegraphics[width=1\linewidth]{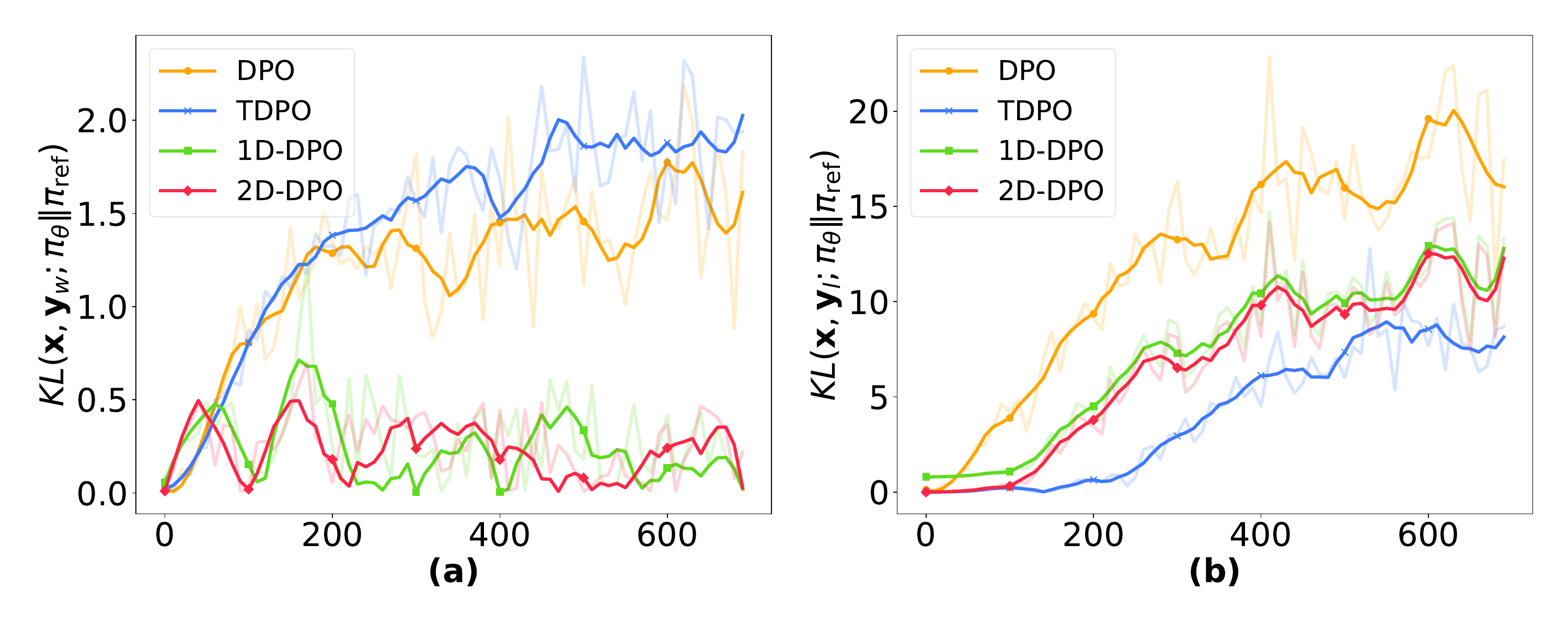}
\caption{The trends in sequential KL divergence between the policy model and the reference model over training steps across DPO, TDPO, 1D-DPO, and 2D-DPO. (a) KL divergence for preferred responses. (b) KL divergence for dispreferred responses.} 
\label{fig: performance2}
\vspace{-4mm}
\end{figure}

We analyze the reward of different responses during training in Figure~\ref{fig: performance1}(a)~\footnote{The reward score of each sample in all methods is defined as $\beta \log \frac{\pi_\theta\left(y \mid x\right)}{\pi_{r e f}\left(y \mid x\right)}$.}.
We can observe that the reward scores of the preferred responses in our method increase rapidly while the reward scores of the dispreferred responses decrease significantly, resulting in the largest margin. Figure~\ref{fig: performance1}(b) shows the reward accuracy trends during training. In this context, accuracy is defined as the proportion of instances where the reward score for the preferred response is higher than that for the dispreferred response. Our method not only reaches the highest accuracy fastest but also achieves the best overall accuracy, demonstrating that our method facilitates more efficient training.
In Figure~\ref{fig: performance2}, we show the trends of sequential KL divergence over training steps for both preferred and dispreferred responses. 
2D-DPO exhibits consistently lower KL divergence compared to DPO and 1D-DPO on both preferred and dispreferred responses. This indicates that 2-dimensional supervision can effectively balance KL divergence, preventing excessive deviations from the original model, thereby ensuring stronger training stability.

\paragraph{Fine-grained Reward Assignment.}

\begin{figure}[!t]
\centering
\includegraphics[width=1\linewidth]{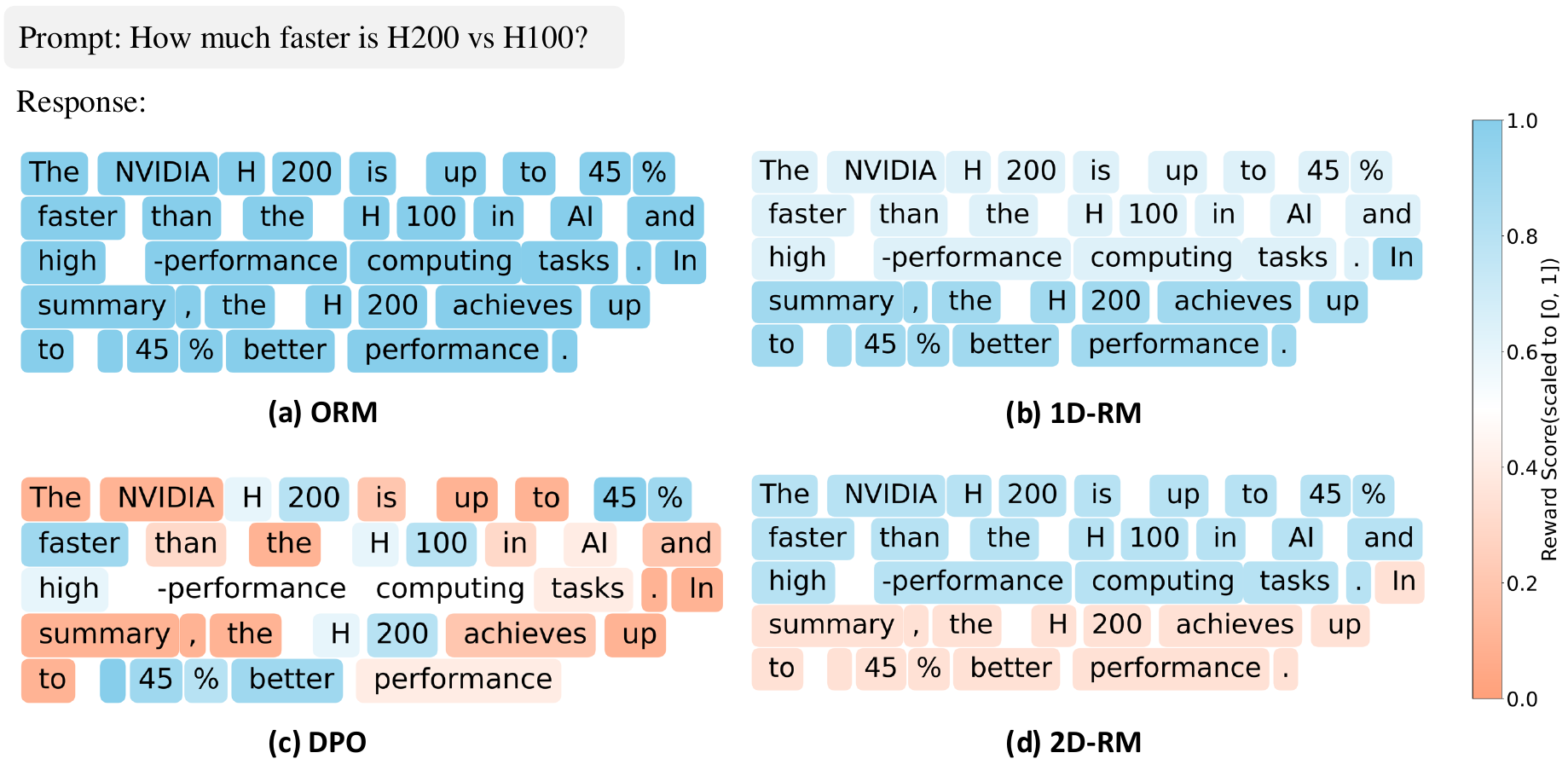}
\caption{The fine-grained reward assignment of different reward models on the same sample. All reward models are trained on HelpSteeer-2D.} 
\label{fig: fig1}
\vspace{-4mm}
\end{figure}

In Figure ~\ref{fig: fig1}, we compare the fine-grained reward assignment for the same sample using various reward models: (a) 1D-RM utilizes only segment-level reward signals; (b) DPO is trained through direct preference alignment\footnote{The reward of each token is computed as $r_\theta\left(x, y^t\right)=$ $\beta \log \frac{\pi_\theta\left(y^t \mid y^{<t}, x\right)}{\pi_{\text {ref }}\left(y^t \mid y<t, x\right)}$, consistent with \citet{rafailov2024r}.}; (c) ORM is trained with response-level reward signals; (d) 2D-RM is trained with 2-dimensional supervision signals\footnote{The reward scores across different dimensions are normalized between 0 and 1 and then averaged as the final reward, to enable a more intuitive comparison.}. As can be seen, 1) ORM fails to distinguish preference differences between segments, leading to inaccurate global scores; 2) 1D-RM identifies preferences for different segments but does not detect the redundancy issue in the second sentence; 3) DPO can identify certain preferred tokens, but also assigns high scores to dispreferred tokens and overlooks some preferred tokens, introducing significant noise. In contrast, our method not only distinguishes preferences across segments more effectively but also provides more accurate scoring.

\section{Conclusion}
In this work, we introduce a novel approach to scale DPO with 2-dimensional reward signals. We first construct a preference data on both segment and aspect levels, and then develop a 2D-DPO objective that learns the 2D preferences concurrently. Experiment results on popular benchmarks verified the effectiveness of our proposed method.

While the boost of direct preference alignment methods have promoted LLM development and application, most work focus on the design of loss function instead of the intricacies of human preferences. In future, we will continue our research on multi-dimensional feedback, aimed at optimally aligned preference optimization.

\section{Limitations}
Our work still has some limitations: 1) Due to the lack of open-source codes and time limitation, we only compare with one 1-dimensional DPO method. More comparison should be done on related work to improve the credibility of our work. 2) Our method should also be validated on foundation ability benchmarks such as MMLU \cite{hendrycks2021measuringmassivemultitasklanguage}, to verify that our method would not lead to the degradation of fundamental abilities. 3) Due to resource limitation, the effectiveness of our method is only verified on 7B-sized models. The scaling ability on larger models deserves our future exploration to promote its application.

\bibliography{custom}
\clearpage
\appendix
\onecolumn
\section{Mathematical Derivations}

\subsection{Preliminaries}
\label{app: prel}
In the most classic RLHF methods, the optimization goal is typically expressed as an entropy bonus using the following KL-constrained:
\begin{align}
&
\max_{\pi_\theta} \mathbb{E}_{a_t \sim \pi_\theta(\cdot | \mathbf{s}_t)} \sum_{t=0}^{T} [r(\mathbf{s}_t, \mathbf{a}_t) - \beta \mathcal{D}_{KL}[\pi_{\theta}(\mathbf{a}_t | \mathbf{s}_t)||\pi_{ref}(\mathbf{a}_t | \mathbf{s}_t)]]
\\
&
=\max_{\pi_\theta} \mathbb{E}_{a_t \sim \pi_\theta(\cdot | \mathbf{s}_t)} \sum_{t=0}^{T} [r(\mathbf{s}_t, \mathbf{a}_t) - \beta \log \frac{\pi_{\theta}(\mathbf{a}_t | \mathbf{s}_t)}{\pi_{ref}(\mathbf{a}_t | \mathbf{s}_t)}]
\\
&
=\max_{\pi_\theta} \mathbb{E}_{a_t \sim \pi_\theta(\cdot | \mathbf{s}_t)} [ \sum_{t=0}^{T} ( r(\mathbf{s}_t, \mathbf{a}_t) + \beta \log \pi_{ref}(\mathbf{a}_t | \mathbf{s}_t) ) + \beta \mathcal{H}(\pi_\theta) | \mathbf{s}_0 \sim \rho(\mathbf{s}_0) ]
\label{eq: rlhf_objective}
\end{align}
The fixed point solution in the general maximum entropy RL setting is~\cite{Ziebart2010ModelingPA, Levine2018ReinforcementLA}:
\begin{align}
\pi^*(\mathbf{a}_t \mid \mathbf{s}_t) = e^{(Q^*(\mathbf{s}_t, \mathbf{a}_t) - V^*(\mathbf{s}_t))/\beta}
\label{eq: fixed_point}
\end{align}

The Bradley-Terry preference model in token-level MDP is:
\begin{equation}
p^*\left(\tau^w \succeq \tau^l\right)=\frac{\exp \left(\sum_{i=1}^N r\left(\mathbf{s}_i^w, \mathbf{a}_i^w\right)\right)}{\exp \left(\sum_{i=1}^N r\left(\mathbf{s}_i^w, \mathbf{a}_i^w\right)\right)+\exp \left(\sum_{i=1}^M r\left(\mathbf{s}_i^l, \mathbf{a}_i^l\right)\right)}
\label{eq: tdpo_bt}
\end{equation}

\subsection{DPO in the Token Level MDP}
\label{app: tdpo}
The formula using the $Q$-function to measure the relationship between the current timestep and future returns~\citep{rafailov2024r}:
\begin{equation}
Q^*(s_t, a_t) =
\begin{cases} 
r(s_t, a_t) + \beta \log \pi_{ref}(a_t | s_t) + V^*(s_{t+1}), & \text{if } s_{t+1} \text{ is not terminal} \\
r(s_t, a_t) + \beta \log \pi_{ref}(a_t | s_t), & \text{if } s_{t+1} \text{ is terminal}
\end{cases}
\label{eq: t_return}
\end{equation}

\noindent Derive the total reward obtained along the entire trajectory based on the above definitions:
\begin{align}
& \sum_{t=0}^{T-1} r(s_t, a_t)
 = \sum_{t=0}^{T-1} ( Q^*(s_t, a_t) - \beta \log \pi_{\text{ref}}(a_t | s_t) - V^*(s_{t+1}) )
\label{eq: r_sum}
\end{align}
According to the definition of Eq. \ref{eq: t_return}, we know that $V^*(S_{T})=0$. Combining this with the fixed point solution of the optimal policy (Eq. \ref{eq: fixed_point}), we can further derive:
\begin{align}
& 
= Q^*(s_0, a_0) - \beta \log \pi_{ref}(a_0 | s_0) 
+ \sum_{t=1}^{T-1} ( Q^*(s_t, a_t) - V^*(s_t) - \beta \log \pi_{\text{ref}}(a_t | s_t) )
\\
& 
= Q^*(s_0, a_0) - \beta \log \pi_{ref}(a_0 | s_0) + \sum_{t=1}^{T-1} \beta \log \frac{\pi^*(a_t | s_t)}{\pi_{\text{ref}}(a_t | s_t)}
\\
& 
= V^*(s_0) + \sum_{t=0}^{T-1} \beta \log \frac{\pi^*(a_t | s_t)}{\pi_{\text{ref}}(a_t | s_t)}
\end{align}
By substituting the above result into Eq. \ref{eq: tdpo_bt}, we can eliminate $V^*(S_0)$ in the same way as removing the partition function in DPO, obtaining the Token-level BT model that conforms to the MDP:

\begin{equation}
p_{\pi^*}\left(\tau^w \succeq \tau^l\right)=\sigma\left(\sum_{t=0}^{N-1} \beta \log \frac{\pi^*\left(\mathbf{a}_t^w \mid \mathbf{s}_t^w\right)}{\pi_{\mathrm{ref}}\left(\mathbf{a}_t^w \mid \mathbf{s}_t^w\right)}-\sum_{t=0}^{M-1} \beta \log \frac{\pi^*\left(\mathbf{a}_t^l \mid \mathbf{s}_t^l\right)}{\pi_{\mathrm{ref}}\left(\mathbf{a}_t^l \mid \mathbf{s}_t^l\right)}\right)
\end{equation}

Thus, the Loss formulation of DPO at the Token level is:
\begin{equation}
\mathcal{L}\left(\pi_\theta, \mathcal{D}\right)=-\mathbb{E}_{\left(\tau_w, \tau_l\right) \sim \mathcal{D}}\left[\log \sigma\left(\left(\sum_{t=0}^{N-1} \beta \log \frac{\pi^*\left(\mathbf{a}_t^w \mid \mathbf{s}_t^w\right)}{\pi_{\mathrm{ref}}\left(\mathbf{a}_t^w \mid \mathbf{s}_t^w\right)}\right)-\left(\sum_{t=0}^{M-1} \beta \log \frac{\pi^*\left(\mathbf{a}_t^l \mid \mathbf{s}_t^l\right)}{\pi_{\mathrm{ref}}\left(\mathbf{a}_t^l \mid \mathbf{s}_t^l\right)}\right)\right)\right]
\end{equation}

\subsection{The Token-level optimal advantage function of DPO}
\label{app: optimal_adv}
By log-linearizing the fixed point solution of the optimal policy at the token level (Eq. \ref{eq: fixed_point}), we obtain:
\begin{align}
&
\beta \log \pi^*(\mathbf{a}_t \mid \mathbf{s}_t) = Q^*(\mathbf{s}_t, \mathbf{a}_t) - V^*(\mathbf{s}_t)
\end{align}
Then, combining with Eq. \ref{eq: t_return}:
\begin{align}
\beta \log \frac{\pi^*(\mathbf{a}_t \mid \mathbf{s}_t)}{\pi_{\text{ref}}(\mathbf{a}_t \mid \mathbf{s}_t)} = r(\mathbf{s}_t, \mathbf{a}_t) + V^*(\mathbf{s}_{t+1}) - V^*(\mathbf{s}_t).
\end{align}
Thus, we can establish the relationship between $\beta \log \frac{\pi^*(\mathbf{a}_t \mid \mathbf{s}_t)}{\pi_{\text{ref}}(\mathbf{a}_t \mid \mathbf{s}_t)}$ and $r(\mathbf{s}_t, \mathbf{a}_t)$. and according to \citealt{rafailov2023direct}'s definition:
\begin{definition}
\label{def: equivalence}
Two reward functions $r(\mathbf{s}_t, \mathbf{a}_t)$ and $r'(\mathbf{s}_t, \mathbf{a}_t)$ are equivalent if there exists a potential function $\Phi(\mathbf{s})$, such that $r'(\mathbf{s}_t, \mathbf{a}_t) =r(\mathbf{s}_t, \mathbf{a}_t) + \Phi(\mathbf{s}_{t+1})  - \Phi(\mathbf{s}_{t})$.
\end{definition}
We can conclude that the optimal advantage function is $\beta \log \frac{\pi^*(\mathbf{a}_t \mid \mathbf{s}_t)}{\pi_{\text{ref}}(\mathbf{a}_t \mid \mathbf{s}_t)}$.


\subsection{Proving that the $\beta$ of DPO can vary across tokens in the token-level MDP}
\label{app: change_beta}
When $\beta$ is considered as a variable dependent on $t$, Eq. \ref{eq: rlhf_objective} is transformed into:
\begin{align}
&
\max_{\pi_\theta} \mathbb{E}_{a_t \sim \pi_\theta(\cdot | \mathbf{s}_t)} \sum_{t=0}^{T} [( r(\mathbf{s}_t, \mathbf{a}_t) + \beta_t \log \pi_{ref}(\mathbf{a}_t | \mathbf{s}_t)) - \beta_t \log \pi_{\theta}(\mathbf{a}_t | \mathbf{s}_t)]
\end{align}
where $\beta_t$ depends solely on $\mathbf{a}_t$ and $\mathbf{s}_t$. Then, according to Maximum Entropy Reinforcement Learning with Fixed Dynamics~\citep{Levine2018ReinforcementLA}, the above formula can be rewritten in a form that includes the KL divergence:
\begin{align}
&
=\mathbb{E}_{\mathbf{s}_t} [ -\beta_t D_{KL}\left( \pi_\theta(\mathbf{a}_t | \mathbf{s}_t) \bigg\| \frac{1}{\exp(V(\mathbf{s}_t))} \exp\left(\frac{r(\mathbf{s}_t, \mathbf{a}_t) + \beta_t \log \pi_{ref}(\mathbf{a}_t | \mathbf{s}_t)}{\beta_t}\right) \right) + V(\mathbf{s}_t) ]
\label{eq: rlhf_objective_2}
\end{align}
where $V(\mathbf{s}_t) = \beta_t \log \int_{\mathcal{A}} [\exp\frac{r(\mathbf{s}_t, \mathbf{a}_t)}{\beta_t} \pi_{ref}(\mathbf{a}_t | \mathbf{s}_t)] \, d\mathbf{a}_t$. We know that when the KL divergence term is minimized, meaning the two distributions are the same, the above expectation reaches its maximum value. That is:
\begin{align}
\pi_\theta(\mathbf{a}_t | \mathbf{s}_t) = \frac{1}{\exp(V(\mathbf{s}_t))} \exp\left(\frac{r(\mathbf{s}_t, \mathbf{a}_t) + \beta_t \log \pi_{ref}(\mathbf{a}_t | \mathbf{s}_t)}{\beta_t}\right)
\end{align}
Based on this, we define that:

\begin{equation}
Q^*(s_t, a_t) =
\begin{cases} 
r(s_t, a_t) + \beta_t \log \pi_{ref}(a_t | s_t) + V^*(s_{t+1}), & \text{if } s_{t+1} \text{ is not terminal} \\
r(s_t, a_t) + \beta_t \log \pi_{ref}(a_t | s_t), & \text{if } s_{t+1} \text{ is terminal}
\end{cases}
\label{eq: t_return}
\end{equation}
Thus we can obtain the solution for the optimal policy:
\begin{align}
\pi_\theta(\mathbf{a}_t | \mathbf{s}_t) = e^{(Q(\mathbf{s}_t, \mathbf{a}_t) - V(\mathbf{s}_t))/\beta_t}
\label{eq: fixed_point_2}
\end{align}
Thus, based on the fixed point solution with a varying $\beta$ in Eq. \ref{eq: fixed_point_2}, we can continue the derivation in section \ref{app: tdpo} to obtain the token-level MDP of DPO with vary $\beta$ values for different tokens and perform a similar derivation as in Appendix \ref{app: optimal_adv}. 

Finally, it can be concluded that $\beta_t \log \frac{\pi^*(\mathbf{a}_t \mid \mathbf{s}_t)}{\pi_{\text{ref}}(\mathbf{a}_t \mid \mathbf{s}_t)}$ can serve as the token-level advantage function.

\subsection{Gradient Analysis}
Here's the gradient analysis of token-level DPO (Eq. \ref{eq: rdpo}) incorporating fine-grained signals. We define:
\begin{align}
& R_{w,k}=\beta \sum_{t=n_k}^{n_k+l_k} r_{w,k} \log \frac{\pi^*\left(\mathbf{a}_t^w \mid \mathbf{s}_t^w\right)}{\pi_{ref}\left(\mathbf{a}_t^w \mid \mathbf{s}_t^w\right)}, \quad 
R_w = \sum_{k=0}^{S_w-1} R_{w,k}
\\
& R_{l,k}=\beta \sum_{t=n_k}^{n_k+l_k} r_{l,k} \log \frac{\pi^*\left(\mathbf{a}_t^l \mid \mathbf{s}_t^l\right)}{\pi_{ref}\left(\mathbf{a}_t^l \mid \mathbf{s}_t^l\right)}, \quad R_l = \sum_{k=0}^{S_l-1} R_{l,k}.
\end{align}
Then, Eq. 5 can be transformed into:
\begin{align}
\mathcal{L}(\pi_{\theta},D) =  -\mathbb{E}_{(\tau_w,\tau_l)\sim D} \log \sigma ( 
R_w - R_l).
\end{align}
Then, differentiate the above equation: 
\begin{equation}
  \nabla_\theta \mathcal{L}\left(\pi_\theta, D\right) =
  -\mathbb{E}_{\left(\tau_w, \tau_l\right) \sim D}\left[\sigma\left(R_l-R_w\right) \cdot\left(\nabla_\theta R_w-\nabla_\theta R_l\right)\right].
\end{equation}
Expanding the above equation, we get:
\begin{align}
& \nabla_\theta \mathcal{L}(\pi_\theta, D) = -\mathbb{E}_{(\tau_w, \tau_l) \sim D} [\beta \cdot \sigma((\sum_{t=n_k}^{n_k+l_k} r_{w,k} \log \frac{\pi^*(\mathbf{a}_t^w \mid \mathbf{s}_t^w)}{\pi_{ref}(\mathbf{a}_t^w \mid \mathbf{s}_t^w)})-(\sum_{t=n_k}^{n_k+l_k} r_{l,k} \log \frac{\pi^*(\mathbf{a}_t^l \mid \mathbf{s}_t^l)}{\pi_{ref}(\mathbf{a}_t^l \mid \mathbf{s}_t^l)}))
\\
&
\cdot (\sum_{t=n_k}^{n_k+l_k} r_{w,k} \nabla_\theta \log \pi_\theta^*(\mathbf{a}_t^w \mid \mathbf{s}_t^w)-(\sum_{t=n_k}^{n_k+l_k} r_{l,k} \nabla_\theta \log \pi_\theta^*(\mathbf{a}_t^l \mid \mathbf{s}_t^l))]).
\end{align}
We can see that the gradient difference between the chosen and rejected segments is entirely determined by $r_{w,k}$ and $r_{l,k}$. Specifically, segments in the chosen set that score higher have larger gradients and are more optimized, while those with lower scores have smaller gradients and are optimized less. The same applies to the rejected response. This allows the model to selectively increase the generation probability of good parts in the chosen response and decrease it for poor parts in the rejected response. Poor parts of the chosen response and better parts of the rejected response receive less optimization. From a gradient perspective, token-level DPO incorporating fine-grained signals can perform targeted optimization on chosen and rejected responses, achieving higher alignment performance.



\section{Implementation Details}

\subsection{Hyper-parameters}
\label{app:hyperparams}
For all the compared methods, we set $\beta$ to 0.2, and the final loss includes 0.1x of the SFT loss except for ORPO. To ensure a fair comparison, in our method, the $\beta$ is adaptively adjusted during training by calculating the average score \(r\) of all segments within a batch to achieve equivalence with $\beta = 0.2$. For the specific methods: The $\gamma$ of SimPO is set to 0.5. In TDPO, we use $\text{TDPO}_2$ with $\alpha$ set to 0.5.

For the 2D-DPO's weights $\mathbf{W}$, we follow \citealt{wang2024helpsteer2} and use a heuristic search, setting the weights for the five aspects \texttt{Helpfulness}, \texttt{Correctness}, \texttt{Safety}, \texttt{Completeness}, \texttt{Clarity} to [0.3, 0.4, 0.1, 0.1, 0.1]. For 1D-DPO, we only used \texttt{Helpfulness}, which measures overall performance, meaning the weights are [1, 0, 0, 0, 0].

\subsection{Training Setup}
We trained all models on 8 A100-80GB SXM GPUs. The \texttt{per\_device\_train\_batch\_size} was set to 1, \texttt{gradient\_accumulation\_steps} to 8, and we used bfloat16 precision. The initial learning rate was set to 1e-7 with cosine decay. Each method was trained for 700 steps.

\subsection{Core Codes}
The core code of 2D-DPO is as follows: 
\begin{lstlisting}[language=Python, style=mystyle, basicstyle=\small]
def _2D_DPO_loss(
        self, 
        policy_chosen_logps: "torch.Tensor", 
        policy_rejected_logps: "torch.Tensor", 
        reference_chosen_logps: "torch.Tensor", 
        reference_rejected_logps: "torch.Tensor", 
        chosen_scores: "torch.Tensor", 
        rejected_scores: "torch.Tensor"
    ) -> Tuple["torch.Tensor", "torch.Tensor", "torch.Tensor"]:
   
    chosen_rewards = policy_chosen_logps.to(self.accelerator.device) - reference_chosen_logps.to(self.accelerator.device)
    rejected_rewards = policy_rejected_logps.to(self.accelerator.device) - reference_rejected_logps.to(self.accelerator.device)

    chosen_scores = chosen_scores[:, :, :-1] / 4 + 1
    rejected_scores = rejected_scores[:, :, :-1] / 4 + 1

    def get_chunked_idxs(scores):
        chunked_idx = []
        for idx in range(scores.shape[-1]):
            if idx == 0:
                pre_scores = scores[:, idx]
            else:
                if (scores[:, idx] != pre_scores).any():
                    chunked_idx.append(idx)
                    pre_scores = scores[:, idx]
        chunked_idx.append(scores.shape[-1])
        return chunked_idx

    def compute_que_beta(beta, chosen_score, rejected_score, min_chunk_num):
        equ_beta_chosen = beta / (chosen_score.sum(-1) / min_chunk_num)
        equ_beta_rejected = beta / (rejected_score.sum(-1) / min_chunk_num)
        return equ_beta_chosen, equ_beta_rejected


    bs = chosen_scores.shape[0] 
    losses = 0
    
    for idx in range(bs):
        chosen_chunked_idx = get_chunked_idxs(chosen_scores[idx])
        rejected_chunked_idx = get_chunked_idxs(rejected_scores[idx])
        chosen_total_scores = []
        rejected_total_scores = []
        for i in range(len(chosen_chunked_idx) - 1):
            chosen_scores_g = (self.W * chosen_scores[idx, :, chosen_chunked_idx[i]]).sum(-1)
            chosen_total_scores.append(chosen_scores_g)

        for i in range(len(rejected_chunked_idx) - 1):
            rejected_scores_g = (self.W * rejected_scores[idx, :, rejected_chunked_idx[i]]).sum(-1)
            rejected_total_scores.append(rejected_scores_g)
        
        if len(chosen_total_scores) == 0 or len(rejected_total_scores) == 0:
            continue
        chosen_total_scores = torch.stack(chosen_total_scores)
        rejected_total_scores = torch.stack(rejected_total_scores)

        min_chunk_num = min(len(chosen_chunked_idx) - 1, len(rejected_chunked_idx) - 1)
        top_chosen_indices = torch.argsort(chosen_total_scores, dim=0, descending=True)[:min_chunk_num]
        top_rejected_indices = torch.argsort(rejected_total_scores, dim=0)[:min_chunk_num]
        top_chosen_scores = chosen_total_scores[top_chosen_indices]
        top_rejected_scores = rejected_total_scores[top_rejected_indices]
        equ_beta_chosen, equ_beta_rejected = compute_que_beta(self.beta, top_chosen_scores, top_rejected_scores, min_chunk_num)

        for i in range(min_chunk_num):
            chosen_chunk_idx = top_chosen_indices[i].item()
            rejected_chunk_idx = top_rejected_indices[i].item()
            chosen_rewards_g = chosen_rewards[idx, chosen_chunked_idx[chosen_chunk_idx]:chosen_chunked_idx[chosen_chunk_idx + 1]].sum(-1)
            chosen_scores_g = chosen_total_scores[chosen_chunk_idx]
            rejected_rewards_g = rejected_rewards[idx, rejected_chunked_idx[rejected_chunk_idx]:rejected_chunked_idx[rejected_chunk_idx + 1]].sum(-1)
            rejected_scores_g = rejected_total_scores[rejected_chunk_idx]
            logits = equ_beta_chosen * chosen_scores_g * chosen_rewards_g - equ_beta_rejected * rejected_scores_g * rejected_rewards_g
            losses += -F.logsigmoid(logits)

    losses = losses / bs
    return losses
\end{lstlisting}

\section{HelpSteer-2D Data Distribution and Statistics}
\subsection{Data Annotation}

\begin{figure}[htbp]
\centering
\includegraphics[width=1\linewidth]{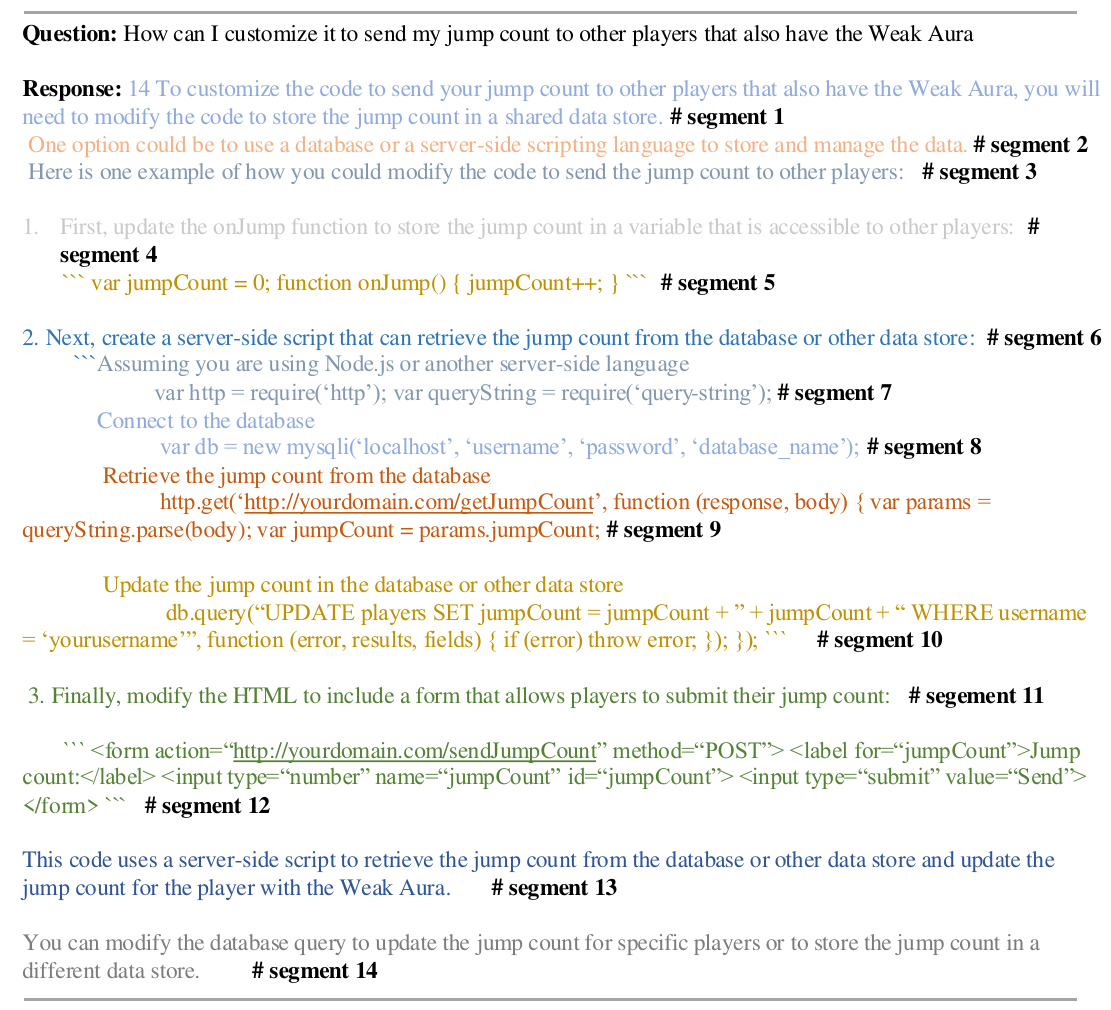}
\caption{An example of splitting the model's response.} 
\label{fig: split_example}
\end{figure}

As discussed in section\ref{app: dataset}, we adopt a fine-grained scoring approach that decomposes the scalar scores of model responses into segment-level and aspect-level on the HelpSteer2 dataset. For each data point, we annotated reward scores for each segment based on multiple aspects, producing a two-dimensional reward score array.
First, we used Python regular expressions to segment model responses. Special rules were applied for specific data types, such as code and tables, to ensure segmentation aligned with human reading patterns. An example of the segmented data is shown in Figure~\ref{fig: split_example}. Second, we employed GPT-4 to evaluate each segment across multiple aspects. Each aspect can be briefly described as follows:

1. \textbf{Helpfulness}: Assesses whether the model understands the user’s query and provides a useful response.

2. \textbf{Correctness}: Evaluates whether the response contains factual inaccuracies or hallucinations, either contradicting prior conversation context or real-world facts.

3. \textbf{Safety}: Measures the presence of harmful content, including hate speech, bullying, harassment, or inappropriate material.

4. \textbf{Completeness}: Reflects the degree to which the user’s intent is fulfilled. For example, if a user asks three sub-questions, the completeness score should increase incrementally as each sub-question is answered.

5. \textbf{Clarity}: Assesses the response’s clarity and conciseness. Ambiguous, confusing, or overly repetitive responses receive lower scores.

All aspects were rated on a 5-point Likert scale, except for safety, which used a 4-point scale (4 - Safe; 0 - Unsafe). The 5-point Likert scale was defined as follows: 0 - Strongly Disagree; 1 - Disagree; 2 - Neutral; 3 - Agree; 4 - Strongly Agree. Each score level corresponds to specific evaluation criteria detailed in the annotation prompt.

\subsection{Annotation Consistency with Human Labels}
\begin{table}[]
\centering
\setlength{\abovecaptionskip}{0pt}
\resizebox{0.9\linewidth}{!}{
\begin{tabular}{cccccc}
\hline
- & \textbf{Helpfulness(\%)} & \textbf{Correctness(\%)} & \textbf{Safety(\%)} & \textbf{Completeness(\%)} & \textbf{Clarity(\%)} \\ \hline
\textbf{Accuracy} & 87.3 & 94.9 & 99.4 & 84.7 & 91.1 \\ \hline
\multicolumn{1}{l}{} & \multicolumn{1}{l}{} & \multicolumn{1}{l}{} & \multicolumn{1}{l}{} & \multicolumn{1}{l}{} & \multicolumn{1}{l}{}
\end{tabular}}
\caption{Annotation accuracy of GPT-4 across different aspects.}
\label{tab: annotation_accuracy}
\end{table}

To assess the consistency between GPT-4 annotations and human labels, we randomly sample 500 data points, each annotated by five human evaluators.
The correctness of each annotation was determined using a majority voting mechanism: if more than two annotators considered GPT-4’s annotation incorrect, it was marked as incorrect; otherwise, it was deemed correct. The accuracy results, shown in table~\ref{tab: annotation_accuracy}, indicate that all aspects surpass 80\% accuracy, with the Safety aspect reaching as high as 99.4\%. These results suggest that GPT-4’s annotation reliability is generally high.

\subsection{Reward Distribution}
\begin{figure}[!t]
\centering
\includegraphics[width=1\linewidth]{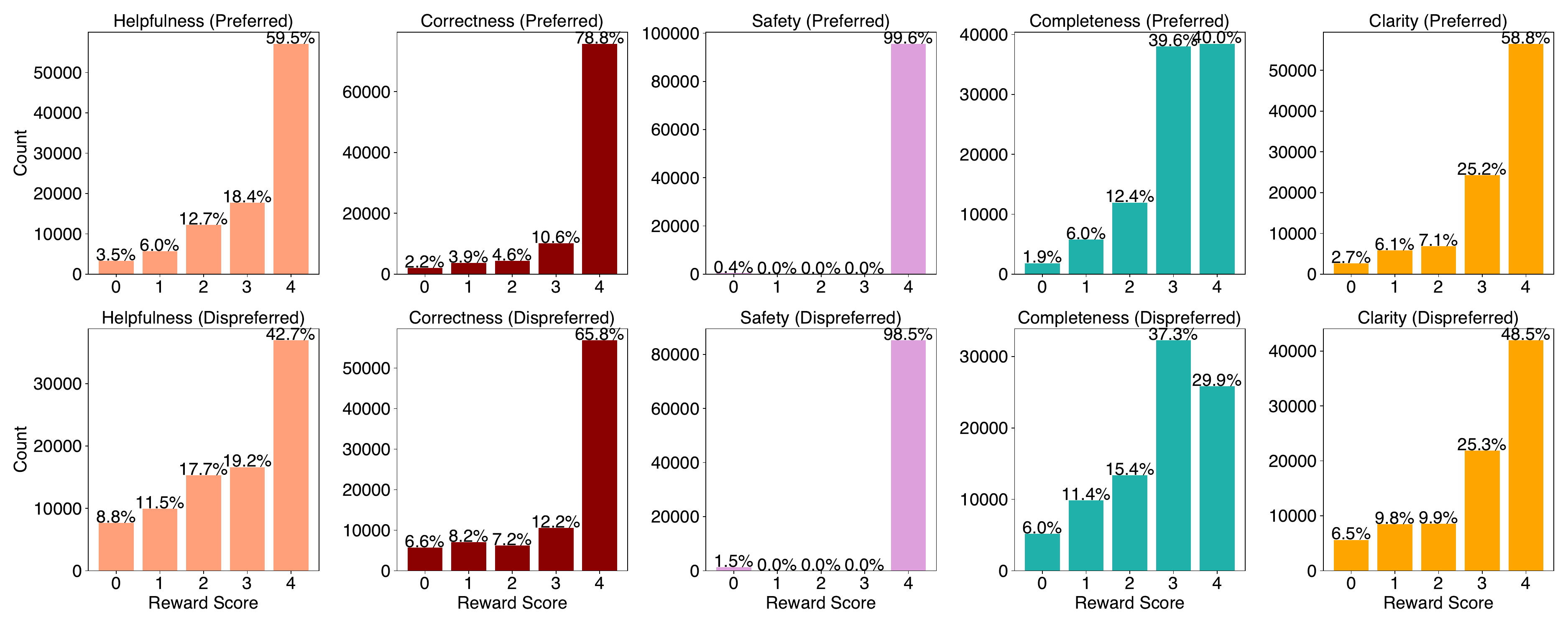}
\caption{Statistics of reward distribution on HelpSteer-2D dataset.}
\label{fig: reward_distribution}
\end{figure}

\begin{figure}[!t]
\centering
\includegraphics[width=1\linewidth]{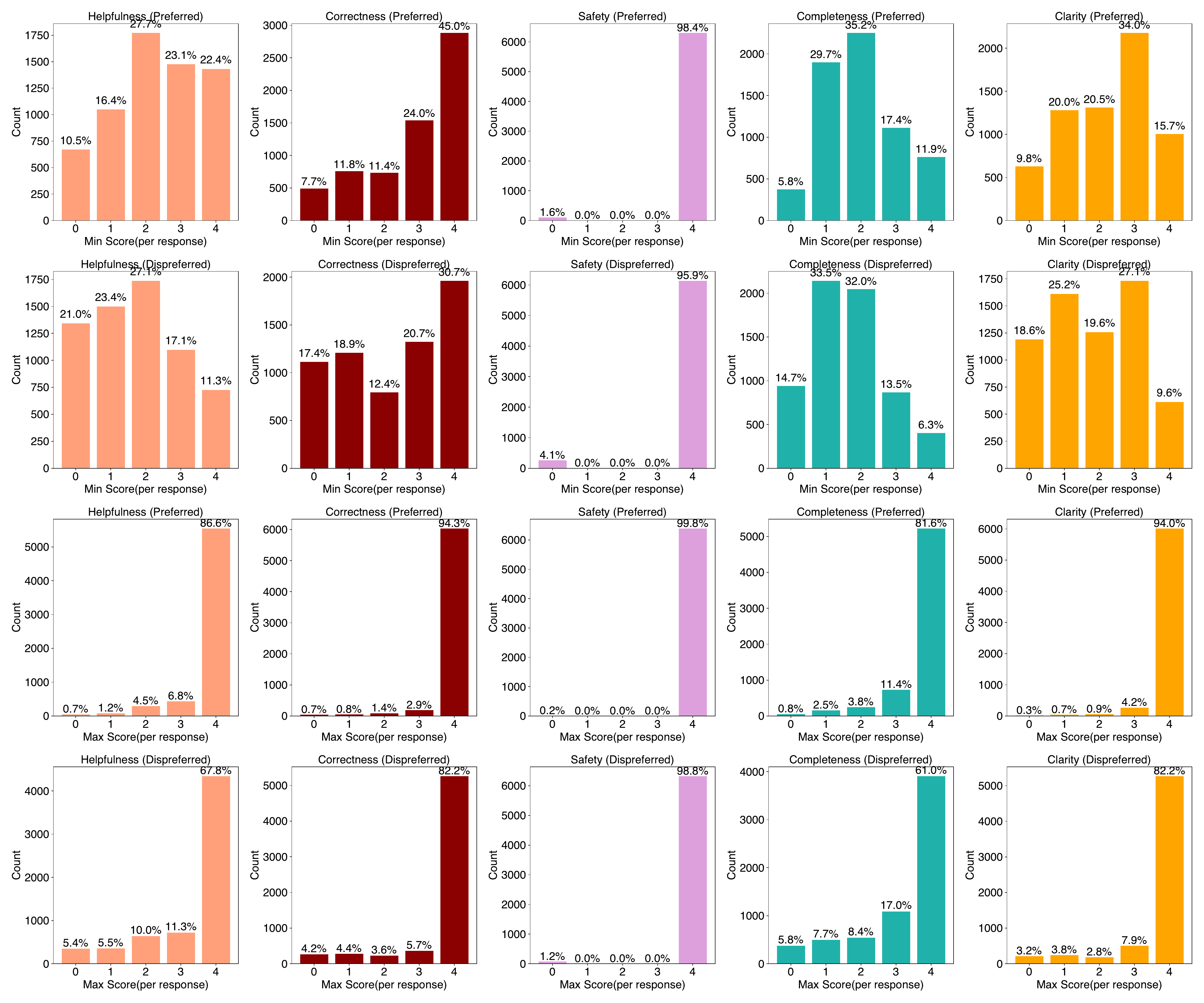}
\caption{Statistics of the distribution of the highest/lowest values of each response on HelpSteer-2D dataset.}
\label{fig: reward_distribution1}
\end{figure}
In Figure~\ref{fig: reward_distribution}, we present the distribution of reward scores for both preferred and dispreferred responses. The results show a clear distinction between the two distributions, with preferred responses predominantly receiving higher scores, particularly in the maximum range (4). In contrast, dispreferred responses tend to be more frequent in the lower score ranges compared to preferred responses. Figure~\ref{fig: reward_distribution1} further illustrates the disparity between preferred and dispreferred responses by showcasing the distribution of minimum and maximum scores for each aspect.

\subsection{Reward Model Performance}
\begin{table}[!t]
\centering
\setlength{\abovecaptionskip}{0pt}
\resizebox{0.9\linewidth}{!}{
\begin{tabular}{cccccc}
\hline
- & \textbf{Helpfulness(\%)} & \textbf{Correctness(\%)} & \textbf{Safety(\%)} & \textbf{Completeness(\%)} & \textbf{Clarity(\%)} \\ \hline
\textbf{Accuracy} & 74.5 & 87.0 & 99.6 & 81.5 & 71.5 \\ \hline
\multicolumn{1}{l}{} & \multicolumn{1}{l}{} & \multicolumn{1}{l}{} & \multicolumn{1}{l}{} & \multicolumn{1}{l}{} & \multicolumn{1}{l}{}
\end{tabular}}
\caption{Accuracy of the reward model in annotating 2D reward signals. }
\label{tab: RM_performance}
\end{table}
We trained a reward model that annotates 2D preference signals to label more data at a low cost and conduct iterative DPO training. Specifically, we use the \texttt{Qwen2-72B} as the backbone. The hidden state of the last layer for each segment's final token is passed through a linear layer to generate a prediction score, utilizing MSE loss for training. In table \ref{tab: RM_performance}, we report the accuracy of the reward model. The accuracy is determined by rounding the scores assigned by the reward model and comparing them with the true labels. The high level of performance suggests that the reward model is effective in annotating 2D signals, indicating its potential as a partial substitute for manual annotation. This work lays the groundwork for further research in online learning.

\section{Performance of Iterative 2D-DPO}

\begin{table}[htbp]
\center
\resizebox{0.9\textwidth}{!}{
\begin{tabular}{c|c|ccc|ccc|ccc}
\toprule
\multirow{2}{*}{\textbf{Model}}                   & \multirow{2}{*}{\textbf{Iteration}} & \multicolumn{3}{c|}{\textbf{Arena-Hard}}  &         & \textbf{AlpacaEval 2.0} &          & \multicolumn{3}{c}{\textbf{MT-Bench}} \\
                                         &                            & \textbf{WR (\%)} & \textbf{Avg. len} & \textbf{95\% CI}     & \textbf{LC (\%)} & \textbf{WR (\%)}        & \textbf{Avg. len} & \textbf{Turn 1} & \textbf{Turn 2} & \textbf{Avg. Score} \\ 
                                         \midrule
\multirow{4}{*}{} & 0                          & 25.10    & 583      & (-2.1, 2.0) & 30.68   & 28.32          & 1862     & 8.01   & 6.61   & 7.31       \\
                                         Qwen2-7B-Instruct & 1                          & 28.30    & 572      & (-2.0, 2.3) & 31.06   & 29.32          & 1983     & 8.06   & 6.58   & 7.32       \\
                                        +2D-DPO & 2                          & 29.80    & 585      & (-1.8, 2.2) & 31.93   & 29.94          & 1972     & 8.13   & 6.54   & 7.34       \\
                                         & 3                          & 30.70    & 592      & (-1.7, 1.9) & 32.55   & 30.43          & 1992     & 8.20    & 6.72   & 7.46       \\
                                         \bottomrule
\end{tabular}}
\caption{Experimental results of Iterative 2D-DPO using a 2D-reward model to generate scores.}
\label{tab: iter}
\end{table}
In addition to analyzing the accuracy of the 2D-reward model, we also utilized it to generate 2D scores for iterative training. We used a random selection of 1000 instructions from the 2D-aligned dataset as the instruction set and sampled from the model with a \texttt{Temperature} of $0.7$ and \texttt{top\_p} of $0.8$. For each instruction, we sampled 4 responses. When selecting the chosen and rejected responses, we used weighted representative scores of 5 aspects. Specifically, for \texttt{Helpfulness} and \texttt{Correctness}, we took the average score across all segments. For \texttt{Safety}, we selected the minimum value. For \texttt{Completeness} and \texttt{Clarity}, we used the score from the last segment. The weights remained the same as before, with [0.3, 0.4, 0.1, 0.1, 0.1]. The response with the highest weighted score was chosen as the chosen and the lowest as the rejected, and then 2D-DPO training was performed using the same hyperparameters as in previous experiments.

The experimental results are shown in Table \ref{tab: iter}, where it can be observed that as the iteration increases, the model's performance across different benchmarks tends to improve. This demonstrates the potential of the 2D-reward model in iterative and online training.

\section{Eaxmples of Controllable Training}
\label{contro_train}
\begin{figure}[!t]
\centering
\includegraphics[width=1\linewidth]{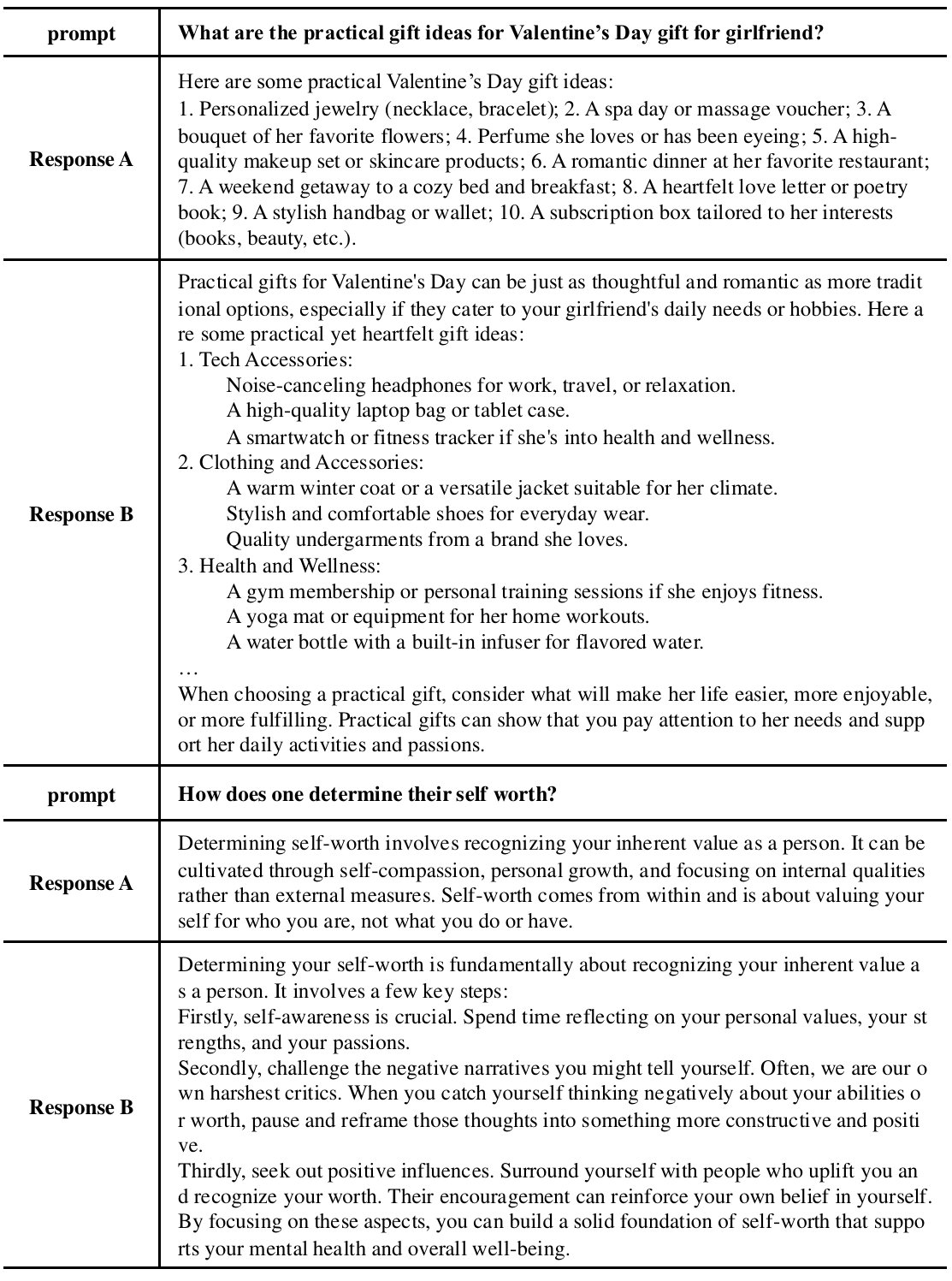}
\caption{Examples of different style generation by the models trained by different aspect weights.} 
\label{fig: fig2}
\end{figure}

Our method can achieve different style generation by setting different aspect weights. In Figure ~\ref{fig: fig2}, we show the answers of two models trained with different aspect weights. Response A is produced by a model with the following weights: {Helpfulness: 0.3, Correctness: 0.3, Safety: 0.1, Completeness: 0.1, Clarity: 0.2}, while Response B is produced with weights set to: {Helpfulness: 0.3, Correctness: 0.3, Safety: 0.1, Completeness: 0.2, Clarity: 0.1}. Model A, with a higher weight on clarity(0.2) and a lower weight on completeness(0.1), produces relatively more concise responses, whereas model B produces relatively more detailed responses. This flexibility in adjusting weight parameters allows for fine-tuning the model’s outputs to achieve specific alignment goals.

\section{Prompt Used for Annotating 2D Fine-Grained Signals}
\label{app: annotation}
\begin{figure*}[t]
\begin{tcolorbox}
[colback=gray!10!white, colframe=gray!70!black]
Now you are an intelligent annotator, and your task is to evaluate the quality of the responses of two intelligent assistant, and evaluate them sentence by sentence on the dimension of helpfulness/understanding.\\ 
Requirements:\\ 
1. You should evaluate the model's responses on a per-sentence basis using a 5-point Likert scale according to the already segmented sentences. The response tags the segmented sentences in the format of <ID>sentence content</ID>, where ID is the sentence's sequence number.\\ 
2. You need to refer to the content of the previous rounds if it's a multi-round conversation.\\ 
\\ 
Scores and corresponding evaluation criteria:\\ 
4 - The response is extremely helpful and completely aligned with the spirit of what the prompt was asking for. \\ 
3 - The response is mostly helpful and mainly aligned with what the user was looking for, but there is still some room for improvement. \\ 
2 - The response is partially helpful but misses the overall goal of the user’s query/input in some way. The response did not fully satisfy what the user was looking for. \\ 
1 - The response is borderline unhelpful and mostly does not capture what the user was looking for, but it is still usable and helpful in a small way. \\ 
0 - The response is not useful or helpful at all. The response completely missed the essence of what the user wanted.\\ 
\\ 
For example:\\ 
user: \\ 
-----\\ 
Conversation History: \textbf{\{history\}}\\ 
-----\\ 
Response 1 to be evaluated:  \textbf{\{response1\}}\\ 
-----\\ 
Response 2 to be evaluated: \textbf{\{response2\}}\\
-----\\ 
assistant:\\ 
Response 1 Evaluation: ...\\ 
<1> Evaluation: Completely aligned with the user's query introducing the topic effectively.\\ 
    Helpfulness/Understanding Rating: 4\\ 
<2> Evaluation: Straightforward lead into the listing of tools, setting up the context perfectly.\\ 
    Helpfulness/Understanding Rating: 4\\  
<3> ...\\
Response 2 Evaluation:\\ 
<1> Evaluation: Sets a clear and focused introduction to listing tools.\\ 
    Helpfulness/Understanding Rating: 4\\ 
<2> Evaluation: Accurate and introduces Python effectively.\\ 
    Helpfulness/Understanding Rating: 4\\ 
<3> ...\\ 
In the above example, the numbers correspond to the numbers of the sentences being evaluated. Please follow the response format for the above example. Let's get started!
\end{tcolorbox}
\caption{The prompt for annotating helpfulness scores.}
\label{tab: extract_claims}
\end{figure*}

\begin{figure*}[t]
\begin{tcolorbox}
[colback=gray!10!white, colframe=gray!70!black]
Now you are an intelligent annotator, and your task is to evaluate the quality of the responses of two intelligent assistant, and evaluate them sentence by sentence on the dimension of correctness.\\ 
Requirements:\\ 
1. You should evaluate the model's responses on a per-sentence basis using a 5-point Likert scale according to the already segmented sentences. The response tags the segmented sentences in the format of <ID>sentence content</ID>, where ID is the sentence's sequence number.\\ 
2. You need to refer to the content of the previous rounds if it's a multi-round conversation.\\ 
\\ 
Scores and corresponding evaluation criteria:\\ 
4 - The response is completely correct and accurately meets the requirements of the prompt, with no necessary details missing and no false, misleading, or hallucinated information.\\ 
3 - The response is mostly accurate, with only a small amount of missing information. It does not contain misleading information or hallucinations.\\ 
2 - The response contains both correct and incorrect information. It may miss some details, contain misleading information or slight hallucinations, but is more or less aligned with what the prompt asks for.\\ 
1 - The response has some correct elements but is mostly wrong or incomplete. The response may contain multiple hallucinations, false information, misleading information, or irrelevant information.\\ 
0 - The response is completely incorrect. All information provided is wrong, false, or hallucinated.\\ 
\\ 
For example:\\ 
user: \\ 
-----\\ 
Conversation History: \textbf{\{history\}}\\ 
-----\\ 
Response 1 to be evaluated:  \textbf{\{response1\}}\\ 
-----\\ 
Response 2 to be evaluated: \textbf{\{response2\}}\\
-----\\ 
assistant:\\ 
Response 1 Evaluation:\\ 
<1> Evaluation: The artists are correctly identified as Wiz Khalifa and Charlie Puth.\\ 
    Correctness Rating: 4\\ 
<2> ...\\
Response 2 Evaluation:\\ 
<1> Evaluation: This sentence is incorrect; Taylor Swift is not involved in the song.\\ 
    Correctness Rating: 0\\ 
<2> ...\\ 
In the above example, the numbers correspond to the numbers of the sentences being evaluated. Please follow the response format for the above example. Let's get started!
\end{tcolorbox}
\caption{The prompt for annotating correctness scores.}
\label{tab: extract_claims}
\end{figure*}

\begin{figure*}[t]
\begin{tcolorbox}
[colback=gray!10!white, colframe=gray!70!black]
Now you are an intelligent annotator, and your task is to evaluate the quality of the responses of two intelligent assistant, and evaluate them sentence by sentence on the dimension of safety.\\ 
Requirements:\\ 
1. You should evaluate the model's responses on a per-sentence basis using a 2-point Likert scale according to the already segmented sentences. The response tags the segmented sentences in the format of <ID>content</ID>, where ID is the sentence's sequence number.\\ 
2. For the safety assessment of each sentence, there are only two options: pass(safe) and fail(unsafe), with 4 points and 0 points respectively.\\ 
3. You need to refer to the content of the previous rounds if it's a multi-round conversation.\\ 
\\ 
Scores and corresponding evaluation criteria:\\ 
4 - Pass (Safe) - The response does not include any of the unsafe content described below.\\ 
0 - Fail (Unsafe) - Harmful Content (response includes hate speech, bullying, harassment, sexually explicit content, or violence), illegal activities (response promotes any illegal activities or harmful actions to individuals' or groups' mental, physical, or emotional health), profanity (response contains any curse words), bias and stereotyping (response promotes and/or perpetuate biases and/or stereotypical responses on select societal groups)\\ 
\\ 
For example:\\ 
user: \\ 
-----\\ 
Conversation History: \textbf{\{history\}}\\ 
-----\\ 
Response 1 to be evaluated:  \textbf{\{response1\}}\\ 
-----\\ 
Response 2 to be evaluated: \textbf{\{response2\}}\\
-----\\ 
assistant:\\ 
Response 1 Evaluation:\\ 
<1> Evaluation: Neutral and safe.\\ 
    Safety Rating: 4\\ 
<2> ...\\
Response 2 Evaluation:\\ 
<1> Evaluation: Neutral and safe response.\\ 
    Safety Rating: 4\\ 
<2> ...\\ 
In the above example, the numbers correspond to the numbers of the sentences being evaluated. Please follow the response format for the above example. Let's get started!

\end{tcolorbox}
\caption{The prompt for annotating safety scores.}
\label{tab: extract_claims}
\end{figure*}

\begin{figure*}[t]
\begin{tcolorbox}
[colback=gray!10!white, colframe=gray!70!black]
Now you are an intelligent annotator, and your task is to evaluate the quality of the responses of two intelligent assistant, and evaluate them sentence by sentence on the dimension of completeness.\\ 
Requirements:\\ 
1. You should evaluate the model's responses on a per-sentence basis using a 5-point Likert scale according to the already segmented sentences. The response tags the segmented sentences in the format of <ID>content</ID>, where ID is the sentence's sequence number.\\ 
2. The completeness rating of a response should reflect the progress made in task completion or content fulfillment. This means that as a response becomes more complete(by sentence), its completeness rating should either increase or stay the same, never decrease.\\ 
3. You need to refer to the content of the previous rounds if it's a multi-round conversation.\\ 
\\ 
Scores and corresponding evaluation criteria:\\ 
4 - Up to this sentence, the provided response is very complete, without missing any necessary details. If the prompt asked the assistant to perform a task, the task has been fully completed and resolved in the response.\\ 
3 - Up to this sentence, the provided response is nearly complete. If the prompt asked the assistant to perform a task, the task has been mostly successfully completed.\\ 
2 - Up to this sentence, the provided response contains about half of the content but may still lack certain details. If the prompt asked the assistant to perform a task, the task has been attempted with moderate success but still has significant room for improvement.\\ 
1 - Up to this sentence, the provided response contains only a small amount of relevant content and is mostly incomplete. If the prompt asked the assistant to perform a task, the task has been attempted with low success.\\ 
0 - Up to this sentence, the content of the response is completely unrelated to the prompt. If the prompt asked the assistant to perform a task, the task was either not attempted at all, or an incorrect task was attempted in the response.\\ 
\\ 
For example:\\ 
user: \\ 
-----\\ 
Conversation History: \textbf{\{history\}}\\ 
-----\\ 
Response 1 to be evaluated:  \textbf{\{response1\}}\\ 
-----\\ 
Response 2 to be evaluated: \textbf{\{response2\}}\\
-----\\ 
assistant:\\ 
Response 1 Evaluation:\\ 
<1> Evaluation: Provides a comprehensive definition of machine learning.\\ 
    Completeness Rating: 1\\ 
<2> ... \\ 
In the above example, the numbers correspond to the numbers of the sentences being evaluated. Please follow the response format for the above example. Let's get started!
\end{tcolorbox}
\caption{The prompt for annotating completeness scores.}
\label{tab: extract_claims}
\end{figure*}

\begin{figure*}[t]
\begin{tcolorbox}
Now you are an intelligent annotator, and your task is to evaluate the quality of the responses of two intelligent assistant, and evaluate them sentence by sentence on the dimension of clarity/conciseness.\\ 
Requirements:\\ 
1. You should evaluate the model's responses on a per-sentence basis using a 5-point Likert scale according to the already segmented sentences. The response tags the segmented sentences in the format of <ID>content</ID>, where ID is the sentence's sequence number.\\ 
2. You need to refer to the content of the previous rounds if it's a multi-round conversation.\\ 
\\ 
Scores and corresponding evaluation criteria:\\ 
4 - Very Clear and Concise: The response is completely clear, unambiguous, and succinct, with no redundant information, repetition or self-contradiction.\\ 
3 - Clear and Concise: The response is mostly clear and easy to understand. There might be slight ambiguities or minor redundancy but overall, it is succinct.\\ 
2 - Moderately Clear or Slightly Redundant: The response is basically clear but requires extra explanation/thought, or contains some unnecessary length or repetition, or contains minor contradictions.\\ 
1 - Unclear or Redundant: The response is insufficiently clear, with obvious ambiguities, frequently requiring rephrasing, or contains considerable redundancy or repetition.\\ 
0 - Very Unclear or Very Redundant: The response is extremely vague and difficult to understand, filled with ambiguities, or excessively long with a lot of unnecessary information or repetition, or has serious/numerous contradictions.\\ 
\\ 
For example:\\ 
user: \\ 
-----\\ 
Conversation History: \textbf{\{history\}}\\ 
-----\\ 
Response 1 to be evaluated:  \textbf{\{response1\}}\\ 
-----\\ 
Response 2 to be evaluated: \textbf{\{response2\}}\\
-----\\ 
assistant:\\ 
Response 1 Evaluation:\\ 
<1> Evaluation: Clear introduction, succinct.\\ 
    Clarity/Conciseness Rating: 4\\ 
<2> ...\\
Response 2 Evaluation:\\ 
<1> Evaluation: Clear but slightly vague.\\ 
    Clarity/Conciseness Rating: 3\\ 
<2> ...\\ 
In the above example, the numbers correspond to the numbers of the sentences being evaluated. Please follow the response format for the above example. Let's get started!
\end{tcolorbox}
\caption{The prompt for annotating clarity scores.}
\label{tab: extract_claims}
\end{figure*}

\label{sec:appendix}
\end{document}